\documentclass[utf8]{article} %
\usepackage[a4paper,%
            left=1in,right=1in,%
            ]{geometry}
\usepackage{url,lineno,microtype,subcaption}
\usepackage[onehalfspacing]{setspace}
\usepackage{appendix}
\usepackage{hyperref}
\hypersetup{
  colorlinks   = true,    
  urlcolor     = blue,    
  linkcolor    = red,    
  citecolor    = red      
}
\usepackage[english]{babel}
\usepackage[open]{bookmark}

\usepackage{natbib}
\usepackage{amsthm}
\usepackage{amsmath,amssymb, amstext}
\usepackage{tikz}
\usepackage{mathtools}
\usepackage{yhmath}
\usepackage{tabu}
\usepackage[capitalize, noabbrev]{cleveref}
\usepackage{bm}
\usepackage{tabularx}
\usepackage{ltablex}
\usepackage{xcolor}
\usepackage{array}   %
\usepackage[colaction]{multicol}

\crefname{appsec}{Appendix}{Appendices}
\newcolumntype{L}{>{$}l<{$}} 
\newcommand\multicollinenumbers{%
	\linenumbers
	\def\makeLineNumber{\docolaction{\makeLineNumberLeft}{}{\makeLineNumberRight}}
}

\newcommand{\X}{\mathcal{X}}
\newcommand{\Y}{\mathcal{Y}}

\newcommand{\E}{\mathcal{E}}
\renewcommand{\S}{\mathcal{S}}
\newcommand{\A}{\mathcal{A}}
\newcommand{\M}{\mathcal{M}}

\makeatletter
\newcommand*{\declarecommand}{%
  \@star@or@long\declare@command
}
\newcommand*{\declare@command}[1]{%
  \provide@command{#1}{}%
  \renew@command{#1}%
}
\makeatother
\newcommand{\myrvspace}[1]{\mathcal{#1}}
\newcommand{\myrvestspace}[1]{\hat{\myrvspace{#1}}}
\newcommand{\myrvest}[1]{\hat{#1}}
\newcommand{\myrvestdist}[1]{q_{\myrvest{#1}}}

\newcommand{\myrvparamrvtheta}[1]{{\Theta_{\myrvest{#1}}}}
\newcommand{\myrvparamrvphi}[1]{{\Phi_{\myrvest{#1}}}}
\newcommand{\myrvparamrvxi}[1]{{\Xi_{\myrvest{#1}}}}
\newcommand{\myrvparamtheta}[1]{{\theta_{\myrvest{#1}}}}
\newcommand{\myrvparamphi}[1]{{\phi_{\myrvest{#1}}}}
\newcommand{\myrvparamxi}[1]{{\xi_{\myrvest{#1}}}}

\newcommand{\myrvparamspacetheta}[1]{{\Delta_{\myrvparamrvtheta{#1}}}}
\newcommand{\myrvparamspacephi}[1]{{\Delta_{\myrvparamrvphi{#1}}}}
\newcommand{\myrvparamspacexi}[1]{{\Delta_{\myrvparamrvxi{#1}}}}

\makeatletter
\newcommand*\defrvar[1]{
  \expandafter\declarecommand\csname s#1\endcsname[1][]{\myrvspace{{#1}}}
  \expandafter\declarecommand\csname h#1\endcsname[1][]{\myrvest{{#1}}}
  \expandafter\declarecommand\csname sh#1\endcsname[1][]{\myrvestspace{#1}}
  \expandafter\declarecommand\csname hp#1\endcsname[1][]{\myrvestdist{#1}}
  \expandafter\declarecommand\csname r#1\endcsname[1][]{\myrvestdist{#1}}
  \expandafter\declarecommand\csname Theta#1\endcsname[1][]{{\myrvparamrvtheta{#1}}}
  \expandafter\declarecommand\csname Phi#1\endcsname[1][]{{\myrvparamrvphi{#1}}}  
  \expandafter\declarecommand\csname Xi#1\endcsname[1][]{{\myrvparamrvxi{#1}}} 
  \expandafter\declarecommand\csname theta#1\endcsname[1][]{{\myrvparamtheta{#1}}}
  \expandafter\declarecommand\csname phi#1\endcsname[1][]{{\myrvparamphi{#1}}}
  \expandafter\declarecommand\csname xi#1\endcsname[1][]{{\myrvparamxi{#1}}}
  \expandafter\declarecommand\csname sTheta#1\endcsname[1][]{\myrvparamspacetheta{#1}}
  \expandafter\declarecommand\csname sPhi#1\endcsname[1][]{\myrvparamspacephi{#1}}
  \expandafter\declarecommand\csname sXi#1\endcsname[1][]{\myrvparamspacexi{#1}}  %
}

\newcommand*\defrvars[1]{
  \@for\@i:=#1\do{\expandafter\defrvar\expandafter{\@i}}}

\newcommand*\defsmallcapsrvar[1]{
  \expandafter\declarecommand\csname h#1\endcsname[1][]{\myrvest{#1}}
}

\newcommand*\defsmallcapsrvars[1]{
  \@for\@i:=#1\do{\expandafter\defsmallcapsrvar\expandafter{\@i}}}  
\makeatother

\defrvars{X,Y,Z,E,S,A,M} %
\defsmallcapsrvars{x,y,z,e,s,a,m}
\newcommand{\pt}{{\prec t}}
\newcommand{\pet}{{\preceq t}}
\newcommand{\pett}{{\preceq t+1}}

\newcommand{\allt}{{0:T}}
\newcommand{\allht}{{0:\hT}}
\newcommand{\ptau}{{\prec \tau}}

\newcommand{\thT}{{t:\hT}}

\newcommand{\hTa}{{\hT_a}}
\newcommand{\thTa}{{t:\hTa}}

\newcommand{\hQ}{{\hat{Q}}}

\newcommand{\sTheta}{{\Delta_\Theta}}
\newcommand{\sPhi}{{\Delta_\Phi}}

\newcommand{\sXi}{{\Delta_\Xi}}

\newcommand{\tp}{{t+1}}
\newcommand{\tm}{{t-1}}
\newcommand{\bB}{\text{\textbf{B}}}
\newcommand{\bA}{\text{\textbf{A}}}
\newcommand{\bD}{\text{\textbf{D}}}
\newcommand{\bQ}{\text{\textbf{Q}}}
\newcommand{\bG}{\text{\textbf{G}}}
\newcommand{\bU}{\text{\textbf{U}}}
\newcommand{\bC}{\text{\textbf{C}}}
\newcommand{\bb}{\text{\textbf{b}}}
\newcommand{\ba}{\text{\textbf{a}}}
\newcommand{\bd}{\text{\textbf{d}}}

\newcommand{\bs}{\text{\textbf{s}}}
\newcommand{\bpi}{\boldsymbol{\pi}}
\newcommand{\bbeta}{\boldsymbol{\beta}}
\newcommand{\boldeta}{\boldsymbol{\eta}}

\newcommand{\phise}{{\phi^1}}
\newcommand{\phieae}{{\phi^2}}
\newcommand{\phie}{{\phi^3}}
\newcommand{\phis}{\phi^*}

\newcommand{\Phise}{{\Phi^1}}
\newcommand{\Phieae}{{\Phi^2}}
\newcommand{\Phie}{{\Phi^3}}
\newcommand{\xise}{{\xi^1}}
\newcommand{\xieae}{{\xi^2}}
\newcommand{\xie}{{\xi^3}}
\newcommand{\Xie}{{\Xi^3}}
\newcommand{\Xise}{{\Xi^1}}
\newcommand{\Xieae}{{\Xi^2}}
\newcommand{\thetase}{{\theta^1}}
\newcommand{\thetaeae}{{\theta^2}}
\newcommand{\thetae}{{\theta^3}}

\newcommand{\phitm}{\phi^{E_{t-1}}}

\newcommand{\phitau}{{\phi^{E_\tau}}}

\newcommand{\phir}{{\phi^{E_r}}}

\newcommand{\phipt}{\phi^{E_\pt}}

\newcommand{\Phipt}{\Phi^{E_\pt}}

\newcommand{\bTheta}{{\boldsymbol{\Theta}}}
\newcommand{\bThetase}{{\boldsymbol{\Theta}^1}}
\newcommand{\bThetaeae}{{\boldsymbol{\Theta}^2}}
\newcommand{\bThetae}{{\boldsymbol{\Theta}^3}}

\newcommand{\xig}{{\xi^\gamma}}
\newcommand{\phig}{\phi^\gamma}
\newcommand{\phigs}{\phi^{\gamma *}}

\newcommand{\Thetase}{{\Theta^1}}
\newcommand{\Thetaeae}{{\Theta^2}}
\newcommand{\Thetae}{{\Theta^3}}

\newcommand{\phinull}{\phi^{E_0}}
\newcommand{\Phinull}{\Phi^{E_0}}

\newcommand{\Phione}{\Phi^{E_1}}

\newcommand{\hT}{{\hat{T}}}

\newcommand{\mot}{\mathfrak{M}}

\newcommand\ci{\perp\!\!\!\perp}
\DeclareMathOperator{\p}{p}
\DeclareMathOperator{\q}{{q}}

\let\r\relax
\DeclareMathOperator{\r}{r}
\let\d\relax
\DeclareMathOperator{\d}{d}

\newcommand*{\diff}{\mathop{\kern0pt\mathrm{d}}\!{}}
\DeclareMathOperator*{\argmax}{arg\,max}
\DeclareMathOperator*{\argmin}{arg\,min}
\DeclareMathOperator{\I}{I}

\DeclareMathOperator{\HS}{H}
\DeclareMathOperator{\KL}{KL}

\DeclareMathOperator{\F}{\mathcal{F}}

\title{Expanding the Active Inference Landscape: More Intrinsic Motivations in the Perception-Action Loop}

\author{Martin Biehl\,$^{1,}$\footnote{martin@araya.org} \and Christian Guckelsberger\,$^{2}$ \and Christoph Salge\,$^{3,4}$ \and Sim\'on C. Smith\,$^{4,5}$ \and Daniel Polani\,$^{4}$}
\date{$^{1}$Araya Inc., Tokyo, Japan\\
$^{2}$Computational Creativity Group, Department of Computing, Goldsmiths, University of London, London, UK\\
$^{3}$Game Innovation Lab, Department of Computer Science and Engineering, New York University, New York City, NY, USA\\
$^{4}$Sepia Lab, Adaptive Systems Research Group, Department of Computer Science, University of Hertfordshire, Hatfield, UK\\
$^{5}$Institute of Perception, Action and Behaviour, School of Informatics, The University of Edinburgh, UK
}

\begin{document}

\maketitle

\begin{abstract}
Active inference is an ambitious theory that treats perception, inference and action selection of autonomous agents under the heading of a single principle. It suggests biologically plausible explanations for many cognitive phenomena, including consciousness. In active inference, action selection is driven by an objective function that evaluates possible future actions with respect to current, inferred beliefs about the world. Active inference at its core is independent from extrinsic rewards, resulting in a high level of robustness across e.g.\ different environments or agent morphologies. In the literature, paradigms that share this independence have been summarised under the notion of intrinsic motivations. In general and in contrast to active inference, these models of motivation come without a commitment to particular inference and action selection mechanisms. In this article, we study if the inference and action selection machinery of active inference can also be used by alternatives to the originally included intrinsic motivation. The perception-action loop explicitly relates inference and action selection to the environment and agent memory, and is consequently used as foundation for our analysis. We reconstruct the active inference approach, locate the original formulation within, and show how alternative intrinsic motivations can be used while keeping many of the original features intact. Furthermore, we illustrate the connection to universal reinforcement learning by means of our formalism. Active inference research may profit from comparisons of the dynamics induced by alternative intrinsic motivations. Research on intrinsic motivations may profit from an additional way to implement intrinsically motivated agents that also share the biological plausibility of active inference. 

\end{abstract}

\tableofcontents

\section{Introduction}
Active inference \citep{friston_active_2012}, and a range of other formalisms usually referred to as intrinsic motivations \citep{storck_reinforcement_1995,klyubin_empowerment_2005,ay_predictive_2008}, all aim to answer a similar question: “Under minimal assumptions, how should an agent act?”. More practically, they relate to what would be a universal way to generate behaviour for an agent or robot that appropriately deals with its environment, i.e.\ acquires the information needed to act and acts towards an intrinsic goal. To this end, both the free energy principle and intrinsic motivations aim to bridge the gap between giving a biologically plausible explanation for how real organism deal with the problem and providing a formalism that can be implemented in artificial agents. Additionally, they share a range of properties, such as an independence of a priori semantics and being defined purely on the dynamics of the agent environment interaction, i.e. the agent's perception-action loop.

Despite these numerous similarities, as far as we know, there has not been any unified or comparative treatment of those approaches. We believe this is in part due to a lack of an appropriate unifying mathematical framework. To alleviate this, we present a technically complete and comprehensive treatment of active inference, including a decomposition of its perception and action selection modes. Such a decomposition allows us to relate active inference and the inherent motivational principle to other intrinsic motivation paradigms such as empowerment \citep{klyubin_empowerment_2005}, predictive information  \citep{ay_predictive_2008}, and knowledge seeking \citep{storck_reinforcement_1995,orseau_universal_2013}. Furthermore, we are able to clarify the relation to universal reinforcement learning \citep{hutter_universal_2005}. Our treatment is deliberately comprehensive and complete, aiming to be a reference for readers interested in the mathematical fundament. 
A considerable number of articles have been published 
on active inference \citep[e.g.][]{friston_active_2012,friston_active_2015,friston_active_2016,friston_active_learning_2016,friston_active_curiosity_2017,friston_graphical_2017,linson_active_2018}. Active inference defines a procedure for both perception and action of an agent interacting with a partially observable environment. The definition of the method, in contrast to other existing approaches~\citep[e.g.][]{hutter_universal_2005,doshi-velez_bayesian_2015,leike_nonparametric_2016}, does not maintain a clear separation between the inference and the action selection mechanisms, and the objective function. Most approaches for perception and action selection are generally formed of three steps: The first step involves a learning or inference mechanism to update the agent's knowledge about the consequences of its actions. In a second step, these consequences are evaluated with respect to an agent-internal objective function. Finally, the action selection mechanism chooses an action depending on the preceding evaluation.

In active inference, these three elements are entangled. On one hand, there is the main feature of active inference: the combination of knowledge updating and action selection into a single mechanism. This single mechanism is the minimisation of a ``variational free energy'' \citep[p.188]{friston_active_2015}. The ``inference'' part of the name is justified by the formal resemblance of the method to the variational free energy minimisation (also known as evidence lower bound maximisation) used in variational inference. Variational inference is a way to turn Bayesian inference into an optimisation problem which gives rise to an approximate Bayesian inference method \citep{wainwright_graphical_2007}. The ``active'' part is justified by the fact that the output of this minimisation is a probability distribution over actions from which the actions of the agent are then sampled. Behaviour in active inference is thus the result of a variational inference-like process. On the other hand, the function (i.e.\ expected free energy) that induces the objective function in active inference is said to be ``of the same form'' as the variational free energy \citep[p.2673]{friston_active_curiosity_2017} or even to ``follow'' from it \citep[p.10]{friston_active_2016}. This suggests that expected free energy is the only objective function compatible with active inference.  

In summary, perception and action in active inference intertwines four elements: variational approximation, inference, action selection, and an objective function. Besides these formal features, active inference is of particular interest for its claims on biological plausibility and its relationship to the thermodynamics of dissipative systems. According to \citet[Section 3]{friston_active_2012}, active inference is a ``corollary'' to the free energy principle. Therefore, it is claimed, actions must minimise variational free energy to resist the dispersion of states of self-organising systems \citep[see also][]{friston_life_2013, allen2016cognitivism}. Active inference has also been used to reproduce a range of neural phenomena in the human brain \citep{friston_active_2016}, and the overarching free energy principle has been proposed as a ``unified brain theory''~\cite{friston2010free}. Furthermore, the principle has been used in a hierarchical formulation as theoretical underpinning of the predictive processing framework~\cite[][pp. 305-306]{clark2015surfing}, successfully explaining a wide range of cognitive phenomena. Of particular interest for the present special issue, the representation of probabilities in the active inference framework is conjectured to be related to aspects of consciousness~\citep{friston_consciousness_2013,linson_active_2018}. 

These strong connections between active inference and biology, statistical physics, and consciousness research make the method particularly interesting for the design of artificial agents that can interact with- and learn about unknown environments. However, it is currently not clear to which extent active inference allows for modifications. We ask: how far do we have to commit to the precise combination of elements used in the literature, and what becomes interchangeable? 

One target for modifications is the objective function. In situations where the environment does not provide a specific reward signal and the goal of the agent is not directly specified, researchers often choose the objective function from a range of \textit{intrinsic motivations}. The concept of intrinsic motivation was introduced as a psychological concept by \citep{ryan_intrinsic_2000}, and is defined as ``the doing of an activity for its inherent satisfactions rather than for some separable consequence''. The concept helps us to understand one important aspect of consciousness: the assignment of affect to certain experiences, e.g.\ the experience of fun~\citep{Dennett1991-DENCE} when playing a game. Computational approaches to intrinsic motivations \citep{oudeyer2009intrinsic,schmidhuber_formal_2010,santucci_which_2013} can be categorised roughly by the psychological motivations they are imitating, e.g.\ drives to manipulate and explore, the reduction of cognitive dissonance, the achievement of optimal incongruity, and finally motivations for effectance, personal causation, competence and self-determination. Intrinsic motivations have been used to enhance behaviour aimed at extrinsic rewards \citep{sutton_reinforcement_1998}, but their defining characteristic is that they can serve as a goal-independent motivational core for autonomous behaviour generation. This characteristic makes them good candidates for the role of value functions for the design of intelligent systems \citep{pfeifer2005}. We attempt to clarify how to modify active inference to accommodate objective functions based on different intrinsic motivations. This may allow future studies to investigate whether and how altering the objective function affects the biological plausibility of active inference. 

Another target for modification, originating more from a theoretical standpoint, is the variational formulation of active inference. As mentioned above, variational inference formulates Bayesian inference as an optimisation problem; a family of probability distributions is optimised to approximate the direct, non-variational Bayesian solution. Active inference is formulated as an optimisation problem as well. We consequently ask: is active inference the variational formulation of a direct (non-variational) Bayesian solution? Such a direct solution would allow a formally simple formulation of active inference without recourse to optimisation or approximation methods, at the cost of sacrificing tractability in most scenarios.

To explore these questions, we take a step back from the established formalism, gradually extend the active inference framework, and comprehensively reconstruct the version presented in \citet{friston_active_2015}. We disentangle the four components of approximation, inference, action selection, and objective functions that are interwoven in active inference. 

One of our findings, from a formal point of view, is that expected free energy can be replaced by other intrinsic motivations. Our reconstruction of active inference then yields a unified formal framework that can accommodate: 
\begin{itemize}
  \item Direct, non-variational Bayesian inference in combination with standard action selection schemes known from reinforcement learning as well as objective functions induced by intrinsic motivations.
  \item Universal reinforcement learning through a special choice of the environment model and a small modification of the action selection scheme.
  \item Variational inference in place of the direct Bayesian approach.
  \item Active inference in combination with objective functions induced by intrinsic motivations.
\end{itemize}

We believe that our framework can benefit active inference research as a means to compare the dynamics induced by alternative action selection principles. Furthermore, it equips researchers on intrinsic motivations with additional ways for designing agents that share the biological plausibility of active inference. 

Finally, this article contributes to the research topic: Consciousness in Humanoid Robots, in several ways. First, there have been numerous claims on how active inference relates to consciousness or related qualities, which we outlined earlier in the introduction. The most recent work by \citet{linson_active_2018}, also part of this research topic, specifically discusses this relation, particularly in regards to assigning salience. Furthermore, intrinsic motivations (including the free energy principle for this argument) have a range of properties that relate to or are useful to a range of classical approaches recently summarised as as Good Old-Fashioned Artificial Consciousness \citep[GOFAC,][]{10.3389/frobt.2018.00039}. For example, embodied approaches still need some form of value-function or motivation \citep{pfeifer2005}, and benefit from the fact that intrinsic motivations are usually universal yet sensitive in regards to an agent's embodiment. %
The enactive AI framework \citep{froese_enactive_2009}, another candidate for GOFAC, proposes further requirements on how value underlying motivation should be grounded in constitutive autonomy and adaptivity. \cite{guckelsberger2016does} present tentative claims on how empowerment maximisation relates to these requirements in biological systems, and how it could contribute to realising them in artificial ones. %
Finally, the idea of using computational approaches for intrinsic motivation goes back to developmental robotics \citep{oudeyer_intrinsic_2007}, where it is suggested as way to produce a learning and adapting robot, which could offer another road to robot consciousness.
Whether these Good Old-Fashioned approaches will ultimately be successful is an open question, and \cite{10.3389/frobt.2018.00039} asses them rather critically. However, extending active inference to alternative intrinsic motivations in a unified framework allows to combine features of these two approaches. For example it may bring together the neurobiological plausibility of active inference and the constitutive autonomy afforded by empowerment.   

\section{Related Work}
\label{sec:relatedWork}
Our work is largely based on \citet{friston_active_2015} and we adopt the setup and models from it. This means many of our assumptions are due to the original paper. 
Recently, \citet{buckley2017free} have provided an overview of continuous-variable active inference with a focus on the mathematical aspects, rather than the relationship to thermodynamic free energy, biological interpretations or neural correlates. Our work here is in as similar spirit but focuses on the discrete formulation of active inference and how it can be decomposed. As we point out in the text, the case of direct Bayesian inference with separate action selection is strongly related to general reinforcement learning \citep{hutter_universal_2005,leike_nonparametric_2016,aslanides_universal_2017}. This approach also tackles unknown environments with- and in later versions also without externally specified reward in a Bayesian way. Other work focusing on unknown environments with rewards are e.g.\ \citet{ross_model-based_2008} and \cite{doshi-velez_bayesian_2015}. We would like to stress that we do not propose agents using Bayesian or variational inference as competitors to any of the existing methods. Instead, our goal is to provide an unbiased investigation of active inference with a particular focus on extending the inference methods, objective functions and action-selection mechanisms. Furthermore, these agents follow almost completely in a straightforward (if quite involved) way from the model in \citet{friston_active_2015}. A small difference is the extension to parameterisations of environment and sensor dynamics. These parameterisations can be found in \citet{friston_active_2016}. %

We note that work on planning as inference \citep{attias_planning_2003,toussaint_probabilistic_2009,botvinick_planning_2012} is generally related to active inference. In this line of work the probability distribution over actions or action sequences that lead to a given goal specified as a sensor value is inferred. Since active inference also tries to obtain a probability distribution over actions the approaches are related. The formalisation of the goal however differs, at least at first sight. How exactly the two approaches relate is beyond the scope of this publication.

\section{Structure of this Article}

Going forward, we will first outline our mathematical notation in \cref{sec:notation}. We then introduce the perception-action loop, which contains both agent and environment in \cref{sec:paloop}. In \cref{sec:inference} we introduce the model used by \citet{friston_active_2015}. We then show how to obtain beliefs about the consequences of actions via both (direct) Bayesian inference (\cref{sec:binference}) and (approximate) variational inference (\cref{sec:approxpostandvi}). These beliefs are represented in the form of a set of complete posteriors. Such a set is a common object but usually does not play a prominent role in Bayesian inference. Here, it turns out to be a convenient structure for capturing the agent' knowledge and describing intrinsic motivations. Under certain assumptions that we discuss in \cref{sec:urlmodel} the direct Bayesian case specialises to the belief updating of the Bayesian universal reinforcement learning agent of \citet{aslanides_universal_2017}. We then discuss in \cref{sec:aselectandim} how those beliefs (i.e.\ the set of complete posteriors) can induce action-value functions (playing the role of objective functions) via a given intrinsic motivation function. We present standard (i.e.\ non-active inference) ways to select actions based on such action-value functions. Then we look at different instances of intrinsic motivation functions. The first is the ``expected free energy'' of active inference. For this we explicitly show how our formalism produces the original expression in \citet{friston_active_2015}. Looking at the formulations of other intrinsic motivations it becomes clear that the expected free energy relies on expressions quite similar or identical to those that occur in other intrinsic motivations. This suggests that, at least in principle, there is no reason why active inference should only work with expected free energy as an intrinsic motivation.
Finally, in \cref{sec:activeinference} formulate active inference for arbitrary action-value functions which include those induced by intrinsic motivations. Modifying the generative model of \cref{sec:genmodel} and looking at the variational approximation of its posterior comes close but does not correspond to the original active inference of \citet{friston_active_2015}. We explain the additional trick that is needed. 

In the appendices we provide some more detailed calculations as well as notation translation tables (\cref{appendix:translationTables}) from our own to those of \citet{friston_active_2015} and \citet{friston_active_2016}.

\section{Notation}
\label{sec:notation}
We will explain our notation in more detail in the text, but for readers that mostly look at equations we give a short summary. Note that, Appendix \ref{appendix:translationTables} comprises a translation between \citet{friston_active_2015,friston_active_2016} and the present notation.
Mostly, we will denote random variables by upper case letters e.g.\ $X,Y,A,E,M,S,...$ their state spaces by calligraphic upper case letters $\X,\Y,\A,\E,\M,\S...$, specific values of random variables which are elements of the state spaces by lower case letters $x,y,a,e,m,s,...$. An exception to this are random variables that act as parameters of probability distributions. For those, we use upper case Greek letters $\Xi,\Phi,\Theta,...$, for their usually continuous state spaces we use $\sXi,\sTheta,\sPhi,...$ and for specific values the lower case Greek letters $\xi,\phi,\theta,...$. 
In cases where a random variable plays the role of an estimate of another variable $X$, we write the estimate as $\hX$, its state space as $\shX$ and its values as $\hx$. 

We distinguish different types of probability distributions with letters $\p,\q,\r$ and $\d$. Here, $\p$ corresponds to probability distributions describing properties of the physical world including the agent and its environment, $\q$ identifies model probabilities used by the agent internally, $\r$ denotes approximations of such model probabilities which are also internal to the agent, and $\d$ denotes a probability distribution that can be replaced by a $\q$ or a $\r$ distribution. We write conditional probabilities in the usual way, e.g.\ $\p(y|x)$. For a model of this conditional probability parameterised by $\theta$, we write $\q(\hy|\hx,\theta)$. 

\section{Perception-Action Loop}
\label{sec:paloop}

\begin{figure}[ht]
\begin{center}
  \begin{tikzpicture}
    [->,>=stealth,auto,node distance=2cm,
    thick]
    \node (e) [] {$E_1$};
    \node (e') [right of=e] {$E_{2}$};
    \node (s) [below of=e, node distance=1cm] {$S_1$};
    \node (s') [below of=e', node distance=1cm] {$S_{2}$};
    \node (a) [below of=s, node distance=1cm] {$A_1$};
    \node (a') [right of=a] {$A_{2}$};
    \node (m) [below of=a, node distance=1cm] {$M_1$};
    \node (m') [below of=a', node distance=1cm] {$M_{2}$};
    \node (el) [left of=e] {$E_0$};
    \node (er) [right of=e'] {};
    \node (sl) [below of=el, node distance=1cm] {$S_0$};
    \node (sr) [below of=er, node distance=1cm] {};
    \node (al) [below of=sl, node distance=1cm] {};
    \node (ar) [right of=a'] {};
    \node (ml) [below of=al, node distance=1cm] {};
    \node (mr) [below of=ar, node distance=1cm] {};

    \path
      (e) edge node {} (e')
      (e) edge node {} (s)
      (e') edge node {} (s')
      (m) edge node {} (a)
      (m) edge node {} (m')
      (m') edge node {} (a')
      (s) edge node {} (m')

      (a) edge node {} (m')
      (a) edge[bend left=45] node {} (e)   
      (a') edge[bend left=45] node {} (e')

      (el) edge node {} (e)
      (el) edge node {} (sl)

      (sl) edge node {} (m)

      (e') edge[-,dotted] node {} (er)
      (s') edge[-,dotted] node {} (mr)
      (m') edge[-,dotted] node {} (mr)
      (a') edge[-,dotted] node {} (mr) 

      ;
  \end{tikzpicture}
  \caption{First two time steps of the Bayesian network representing the perception-action loop (PA-loop). All subsequent time steps are identical to the one from time $t=1$ to $t=2$.}
  \label{fig:smloop}
\end{center}
\end{figure}
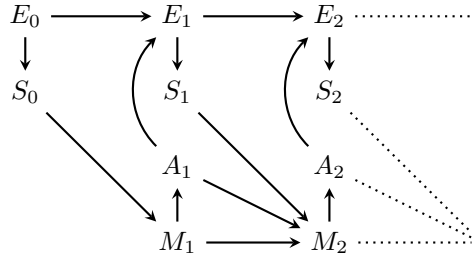

In this section we introduce an agent's perception-action loop (PA-loop) as a causal Bayesian network. This formalism forms the basis for our treatment of active inference. The PA-loop should be seen as specifying the (true) dynamics of the underlying physical system that contains agent and environment as well as their interactions. In Friston's formulation, the environment dynamics of the PA-loop are referred to as the \textit{generative process}. In general these dynamics are inaccessible to the agent itself. 
Nonetheless, parts of these (true) dynamics are often assumed to be known to the agent in order to simplify computation \citep[see e.g.~][]{friston_active_2015}. 
We first formally introduce the PA-loop as causal Bayesian network, and then state specific assumptions for the rest of this article.

\subsection{PA-loop Bayesian Network}
\cref{fig:smloop} shows an agent's PA-loop, formalised as causal Bayesian network. The network describes the following causal dependencies over time:
At $t=0$ an initial environment state $e_0 \in \sE$ leads to an initial sensor value $s_0 \in \sS$. This sensor value influences the memory state $m_1 \in \sM$ of the agent at time $t=1$. Depending on this memory state, action $a_1 \in \sA$ is performed which influences the transition of the environment state from $e_0$ to $e_1 \in \sE$. The new environment state leads to a new sensor value $s_1$ which, together with the performed action $a_1$ and the memory state $m_1$, influence the next memory state $m_2$. The loop then continues in this way until a final time step $T$. 

We assume that all variables are finite and that the PA-loop is time-homogeneous\footnote{This means that all state spaces and transition probabilities are independent of the time step, e.g.\ $\sM_{t}=\sM_{t-1}$ and $p(s_t|e_t)=p(s_{t-1}|e_{t-1})$.}. %
We exclude the first transition from $t=0$ to $t=1$ from the assumption of time-homogeneity in order to avoid having to pick an arbitrary action which precedes the investigated time-frame. The first transition is thus simplified to $\p(m_1|s_0,a_0):= \p(m_1|s_0)$. 
Under the assumption of time-homogeneity and the causal dependencies expressed in \cref{fig:smloop}, the joint probability distribution over the entire PA-loop is defined by:

\begin{align}
  \p(e_{0:T},s_{0:T},a_{1:T},m_{1:T})&= 
 \left( \prod_{t=1}^T \p(a_t|m_t) \p(m_t|s_{t-1},a_{t-1}) \p(s_t|e_t) \p(e_t|a_t,e_{t-1}) \right) \p(s_0|e_0) \p(e_0)
\end{align} 
where $e_{0:T}$ is shorthand for states $(e_0, e_1, ..., e_T)$. In order to completely determine this distribution we therefore have to specify the state spaces $\sE,\sS,\sA$, and $\sM$ as well as the following probabilities and mechanisms for all $e_0,e_t,e_{t+1} \in \sE; s_0,s_t \in \sS; a_t,a_{t+1} \in \sA; m_1,m_t,m_{t+1} \in \sM$ for $t>0$:
\begin{multicols}{2}
	\multicollinenumbers
	\begin{itemize}
		\item initial environment distribution: $\p(e_0)$,
		\item environment dynamics: $\p(e_{t+1}|a_{t+1},e_t)$,
		\item sensor dynamics: $\p(s_t|e_t)$,
		\item action generation: $\p(a_t|m_t)$,
		\item initial memory step $\p(m_1|s_0)$,
		\item memory dynamics: $\p(m_{t+1}|s_t,a_t,m_t)$.
	\end{itemize}
\end{multicols}

In the following we will refer to a combination of initial environment distribution, environment dynamics, and sensor dynamics simply as an \textit{environment}. Similarly, an \textit{agent} is a particular combination of initial memory step, memory dynamics, and action generation. 
The indexing convention we use here is identical to the one used for the generative model (see \cref{sec:genmodel}) in \citet{friston_active_2015}.

Also, note the dependence of $M_t$ on $S_{t-1}$, $M_{t-1}$, and additionally $A_{t-1}$ in \cref{fig:smloop}. 
In the literature, the dependence on $A_{t-1}$ is frequently not allowed \citep{ay_information-driven_2012,ay_umwelt_2015}. However, we assume an “efference”-like update of the memory. Note that this dependence in addition to the dependence on $m_{t-1}$ is only relevant if the actions are not deterministic functions of the memory state\footnote{In the deterministic case there is a function $f:\sM \rightarrow \sA$ such that $p(m_t|s_\tm,a_\tm,m_\tm)=p(m_t|s_\tm,f(m_\tm),m_\tm)=p(m_t|s_\tm,m_\tm)$.}.
If action selection is probabilistic, knowing the outcome $a_{t-1}$ of the action generation mechanism $\p(a_{t-1}|m_{t-1})$ will convey more information than only knowing the past memory state $m_{t-1}$. This additional information can be used in inference about the environment state and fundamentally change the intrinsic perspective of an agent. We do not discuss these changes in more detail here but the reader should be aware of the assumption.

In a realistic robot scenario, the action $a_t$, if it is to be known by the agent, can only refer to the ``action signal'' or ``action value'' that is sent to the robot's physical actuators. These actuators will usually be noisy and the robot will not have access to the final effect of the signal it sends. The (noisy) conversion of an action signal to a physical configuration change of the actuator is here seen as part of the environment dynamics $\p(e_{t}|a_{t},e_{t-1})$. Similarly, the sensor value is the signal that the physical sensor of the robot produces as a result of a usually noisy measurement, so just like the actuator, the conversion of a physical sensor configuration to a sensor value is part of the sensor dynamics $\p(s_t|e_t)$ which in turn belongs to the environment. As we will see later, the actions and sensor values must have well defined state spaces $\sA$ and $\sS$ for inference on an internal model to work. This further justifies this perspective.

\subsection{Assumptions}
\label{sec:paloopassumptions}
For the rest of this article we assume that the environment state space $\sE$, sensor state space $\sS$ as well as environment dynamics $\p(e_{t+1}|a_{t+1},e_t)$ and sensor dynamics $\p(s_t|e_t)$ are arbitrarily fixed and that some initial environmental state $e_0$ is given. Since we are interested in intrinsic motivations, our focus is not on specific environment or sensor dynamics but almost exclusively on action generation mechanisms of agents that rely minimally on the specifics of these dynamics. 

In order to focus on action generation, we assume that all the agents we deal with here have the same memory dynamics. For this, we choose a memory that stores all past sensor values $s_\pt=(s_0,s_1,...,s_\tm)$ and actions $a_\pt=(a_1,a_2,...,a_\tm)$ in the memory state $m_t$. This type of memory is also used in \citet{friston_active_2015,friston_active_2016} and provides the agent with all existing data about its interactions with the environment. In this respect, it could be called a perfect memory. At the same time, whatever the agent learned from $s_\pt$ and $a_\pt$ that remains true based on the next time step's $s_\pett$ and $a_\pett$ must be relearned from scratch by the agent. A more efficient memory use might store only a sufficient statistic of the past data and keep reusable results of computations in memory. Such improvements are not part of this article  \citep[see e.g.][for discussion]{fox_minimum-information_2016}. 

Formally, the state space $\sM$ of the memory is the set of all sequences of sensor values and actions that can occur. Since there is only a sensor value and no action at $t=0$, these sequences always begin with a sensor value followed by pairs of sensor values and actions. Furthermore, the sensor value and action at $t=T$ are never recorded. Since we have assumed a time-homogeneous memory state space $\sM$ we must define it so that it contains all these possible sequences from the start. Formally, we therefore choose the union of the spaces of sequences of a fixed length (similar to a Kleene-closure): 
\begin{equation}
\label{eq:memorystatespace}
\sM=\sS \cup \left(\bigcup_{t=1}^{T-1} \sS\times (\sS \times \sA)^t\right).                                                                                                                                                                                                                                                                                                                                                                                                                                                                                                                                                                                                                                                                                                                                                                                                                                                                 \end{equation}
With this we can define the dynamics of the memory as: 
\begin{align}
\p(m_1|s_0):&= \begin{cases} 1 &\text{ if } m_1 = s_0 \\
                                                  0 &\text{ else.}    
                                    \end{cases}\\
  \p(m_t|s_{t-1},a_{t-1},m_{t-1}) :&= \begin{cases} 1 &\text{ if } m_t = m_\tm s_{t-1} a_{t-1} \\
                                                  0 &\text{ else.}    
                                    \end{cases} 
\end{align} 
This perfect memory may seem unrealistic and can cause problems if the sensor state space is large (e.g. high resolution images). However, we are not concerned with this type of problem here. Usually, the computation of actions based on past actions and sensor values becomes a challenge of efficiency long before storage limitations kick in: the necessary storage space for perfect memory only increases linearly with time, while, as we show later, the number of operations for Bayesian inference increases exponentially.

For completeness we also note how the memory dynamics look if actions are a deterministic function $f:\sM \rightarrow \sA$ of the memory state. Recall that in this case we can drop the edge from $A_{t-1}$ to $M_t$ in the PA-loop in \cref{fig:smloop} and have $a_t=f(m_t)$ so that we can define:
\begin{align}
\p(m_1|s_0):&= \begin{cases} 1 &\text{ if } m_1 = s_0 \\
                                                  0 &\text{ else.}    
                                    \end{cases}\\
  \p(m_t|s_{t-1},m_{t-1}) :&= \begin{cases} 1 &\text{ if } m_t = m_\tm s_{t-1} f(m_{t-1}) \\
                                                  0 &\text{ else.}    
                                    \end{cases} 
\end{align} 
Given a fixed environment and the memory dynamics, we only have to define the action generation mechanism $\p(a_t|m_t)$ to fully specify the perception-action loop. This is the subject of the next two sections. 

In order to stay as close to \citet{friston_active_2015} as possible, we first explain the individual building blocks that can be extracted from Friston's active inference as described in \citet{friston_active_2015}. These are the variational inference and the action selection. We then show how these two building blocks are combined in the original formulation. We eventually leverage our separation of components to show how the action selection component can be modified, and thus extend the active inference framework.

\section{Inference and Complete Posteriors}
\label{sec:inference}
Ultimately, an agent needs to select actions. Inference based on past sensor values and actions is only needed if it is relevant to the action selection. Friston's active inference approach promises to perform action selection within the same inference step that is used to update the agent's model of the environment. In this section, we look at the inference component only and show how an agent can update a generative model in response to observed sensor values and performed actions.

The natural way of updating such a model is Bayesian inference via Bayes' rule. This type of inference leads to what we call the \textit{complete posterior}. The complete posterior represents all knowledge that the agent can obtain about the consequences of its actions from its past sensor values and actions. In \cref{sec:aselectandim} we discuss how the agent can use the complete posterior to decide what is the best action to take. %

Bayesian inference as straightforward recipe is usually not practical due to computational costs. The memory requirements of the complete posterior update increases exponentially with time and so does the number of operations needed to select actions. To keep the computational tractable, we have to limit ourselves to only use parts of the complete posterior. Furthermore, since the direct expressions (even of parts) of complete posteriors are usually intractable, approximations are needed. Friston's active inference is committed to variational inference as an approximation technique. Therefore, we explain how variational inference can be used as an approximation technique. Our setup for variational inference (generative model and approximate posterior) is identical to the one in \citet{friston_active_2015}, but in this section we ignore the inference of actions included there. We will look at the extension to action inference in \cref{sec:aselectandim}.

In the perception-action loop in \cref{fig:smloop}, action selection (and any inference mechanism used in the course of it) depends exclusively on the memory state $m_t$. As mentioned in \cref{sec:paloop}, we assume that this memory state contains all \textit{past} sensor values $s_\pt$ and all \textit{past} actions $a_\pt$. To save space, we write $sa_\pt:=(s_\pt,a_\pt)$ to refer to both sensor values and actions. We then have: 
\begin{equation}
  m_t = sa_\pt.
\end{equation}  
However, since it is more intuitive to understand inference with respect to past sensor values and actions than in terms of memory, we use $sa_\pt$ explicitly here in place of $m_t$.

\subsection{Generative Model}
\label{sec:genmodel}
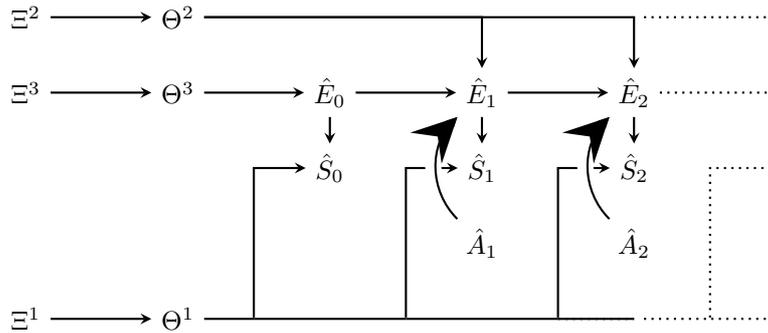
\begin{figure}[h!]%
	\begin{center}
		\begin{tikzpicture}
		[->,>=stealth,auto,node distance=2cm,
		thick]
		\tikzset{
			hv/.style={to path={-| (\tikztotarget)}},
			vh/.style={to path={|- (\tikztotarget)}},
		}
		\tikzset{invi/.style={minimum width=0mm,inner sep=0mm,outer sep=0mm}}

		\node (e) [] {$\hE_1$};
		\node (e') [right of=e] {$\hE_{2}$};
		\node (s) [below of=e, node distance=1cm] {$\hS_1$};
		\node (s') [below of=e', node distance=1cm] {$\hS_{2}$};
		\node (a) [below of=s, node distance=1cm] {$\hA_1$};
		\node (a') [right of=a] {$\hA_{2}$};
		\node (m') [below of=a', node distance=1cm] {};
		\node (el) [left of=e] {$\hE_0$};
		\node (er) [right of=e'] {};
		\node (sl) [below of=el, node distance=1cm] {$\hS_0$};
		\node (sr) [below of=er, node distance=1cm] {};
		\node (al) [below of=sl, node distance=1cm] {};
		\node (ar) [right of=a'] {};
		\node (ml) [below of=al, node distance=1cm] {};
		\node (mr) [below of=ar, node distance=1cm] {};
		\node (th3) [left of=el] {$\Thetae$};
		\node (th2) [above of=th3, node distance=1cm] {$\Thetaeae$};
		\node (th1) [below of=th3, node distance=3cm] {$\Thetase$};
		
		\node (al3) [left of=th3] {$\Xie$};
		\node (al2) [left of=th2] {$\Xieae$};
		\node (al1) [left of=th1] {$\Xise$};

		\node (th2') [above of=e', node distance=1cm] {};  
		\node (th2r) [right of=th2'] {};   
		
		\node (c0) [right of=th1, node distance=1cm,invi] {};
		\node (c1) [right of=c0,invi] {};
		\node (c2) [right of=c1,invi] {};
		\node (c3) [right of=c2,invi] {};
		\node (c3a) [right of=c2,node distance=1cm,invi] {};
		
		\path
		(al3) edge (th3)
		(al2) edge (th2)
		(al1) edge (th1)
		
		(th3) edge (el)
		(th2) edge[hv] (e)
		(th2) edge[hv] (e')
		(th2') edge[-,dotted] (th2r)
		(th1) edge[-] (c0)
		(c0) edge[vh] (sl)
		(c0) edge[-] (c1)
		(c1) edge[vh] (s)
		(c1) edge[-] (c2)
		(c2) edge[vh] (s')
		(c2) edge[-] (c3a)
		(c2) edge[-,dotted] (c3a)
		
		(c3) edge[-,dotted,vh] (sr)
		
		(e) edge node {} (e')
		(e) edge node {} (s)
		(e') edge node {} (s')
		(a) edge[bend left=45,line width=6pt,draw=white] node {} (e)   
		(a') edge[bend left=45,line width=6pt,draw=white] node {} (e')
		
		(a) edge[bend left=45] node {} (e)   
		(a') edge[bend left=45] node {} (e')

		(el) edge node {} (e)
		(el) edge node {} (sl)

		(e') edge[-,dotted] node {} (er)
		(m') edge[-,dotted] node {} (mr)

		;
		\end{tikzpicture}
		\caption{Bayesian network of the generative model with parameters $\bTheta=(\bThetase,\bThetaeae,\bThetae)$ and hyperparameters $\Xi=(\Xise,\Xieae,\Xie)$. Hatted variables are models / estimates of non-hatted counterparts in the perception-action loop in \cref{fig:smloop}. An edge that splits up connecting one node to $n$ nodes (e.g. $\bThetaeae$ to $\hE_1, \hE_2, ...$) corresponds to $n$ edges from that node to all the targets under the usual Bayesian network convention. Note that in contrast to the perception-action loop in \cref{fig:smloop}, imagined actions $\hA_t$ have no parents. They are either set to past values or, for those in the future, a probability distribution over them must be assumed.} 
		\label{fig:genmodel}
	\end{center}
\end{figure}

The inference mechanism, internal to the action selection mechanism $\p(a|m)$, takes place on a hierarchical generative model (or density, in the continuous case). ``Hierarchical'' means that the model has parameters and hyperparameters, and ``generative'' indicates that the model relates \emph{parameters and latent variables}, i.e. the environment state, as ``generative'' causes to sensor values and actions as \emph{data} in a joint distribution. %
The generative model we investigate here is a part of the generative model used in \citet{friston_active_2015}. For now, we omit the probability distribution over future actions and the ``precision'', which are only needed for active inference and are discussed later. The generative models in \citet{friston_active_learning_2016,friston_active_2016,friston_active_curiosity_2017} are all closely related. 

Note that we are not inferring the causal structure of the Bayesian network or state space cardinalities, but define the generative model as a fixed Bayesian network with the graph shown in \cref{fig:genmodel}. It is possible to infer the causal structure \citep[see e.g.][]{ellis_learning_2008}, but in that case, it becomes impossible to represent the whole generative model as a single Bayesian network \citep{ortega_bayesian_2011}. 

The variables in the Bayesian network in \cref{fig:genmodel} that model variables occurring outside of $\p(a|m)$ in the perception-action loop (\cref{fig:smloop}), are denoted as hatted versions of their counterparts. 
More precisely:
\begin{itemize}
  \item $\hs \in \shS=\sS$ are modelled sensor values,  
\item $\ha \in \shA=\sA$ are modelled actions,
\item $\he \in \shE$ are modelled environment states.
\end{itemize}
To clearly distinguish the probabilities defined by the generative model from the true dynamics, we use the symbol $\q$ instead of $\p$. In accordance with \cref{fig:genmodel}, and also assuming time-homogeneity, the joint probability distribution over all variables in the model until some final modelled time $\hT$ is given by:
\begin{align}
\label{eq:genmodel}
\begin{split}
  \q(\he_{0:T},&\hs_{0:T},\ha_{1:T},\thetase,\thetaeae,\thetae,\xise,\xieae,\xie):=\\
  &\left(\prod_{t=1}^T \q(\hs_t|\he_t,\thetase) \q(\he_t|\ha_t,\he_{t-1},\thetaeae) \q(\ha_t) \right) \q(\hs_0|\he_0,\thetase)\q(\he_0|\thetae) \left(\prod_{i=1}^3 \q(\theta^i|\xi^i) \q(\xi^i)\right)
  \end{split}
\end{align} 
Here, $\thetase,\thetaeae,\thetae$ are the parameters of the hierarchical model, and $\xise,\xieae,\xie$ are the hyperparameters. To save space, we combine the parameters and hyperparameters by writing 
\begin{align}
\theta&:=(\thetase,\thetaeae,\thetae)\\
\xi&:=(\xise,\xieae,\xie).
\end{align}
To fully specify the generative model, or equivalently a probability distribution over \cref{fig:genmodel}, we have to specify the state spaces $\shE,\shS,\shA$ and:
\begin{multicols}{2}
\multicollinenumbers
\begin{itemize}
  \item $\q(\hs|\he,\thetase)$ the sensor dynamics model,
  \item $\q(\he'|\ha',\he,\thetaeae)$ the environment dynamics model,
  \item $\q(\he_0|\thetae)$ the initial environment state model,
  \item $\q(\thetase|\xise)$ the sensor dynamics prior,
  \item $\q(\thetaeae|\xieae)$ the environment dynamics prior,
  \item $\q(\thetae|\xie)$ the initial environment state prior,
  \item $\q(\xise)$ sensor dynamics hyperprior,
  \item $\q(\xieae)$ environment dynamics hyperprior,
  \item $\q(\xie)$ initial environment state hyperprior,
  \item $\hT$ last modelled time step,
  \item $\q(\ha_t)$ for all $t \in \{1,,...,\hT\}$ the probability distribution over the actions at time $t$.
\end{itemize}
\end{multicols}
The state spaces of the parameters and hyperparameters are determined by the choice of $\shE,\shS,\shA$. We will see in \cref{sec:plugin} that $\shS=\sS$ and $\shA=\sA$ should be chosen in order to use this model for inference on past sensor values and actions. For $\shE$ it is not necessary to set it equal to $\sE$ for the methods described to work. We note that if we set $\shE$ equal to the memory state space of \cref{eq:memorystatespace} the model and its updates become equivalent to those used by the Bayesian universal reinforcement learning agent  \citet{hutter_universal_2005} in a finite (environment and time-interval) setting (see \cref{sec:urlmodel}). 

The last modelled time step $\hT$ can be chosen as $\hT=T$, but it is also possible to always set it to $\hT=t+n$, in which case $n$ specifies a future time horizon from current time step $t$. Such an agent would model a future that goes beyond the externally specified last time step $T$. The dependence of $\hT$ on $t$ (which we do not denote explicitly) within $\p(a|m)$ is possible since the current time step $t$ is accessible from inspection of the memory state $m_t$ which contains a sensor sequence of length $t$. 

The generative model assumes that the actions are not influenced by any other variables, hence we have to specify action probabilities. This means that the agent does not model how its actions come about, i.e.\ it does not model its own decision process. Instead, the agent is interested in the (parameters of) the environment and sensor dynamics. It actively sets the probability distributions over past and future actions according to its needs. In practice, it either fixes the probability distributions to particular values (by using Dirac delta distributions) or to values that optimise some measure. We look into the optimisation options in more detail later.

Note that the parameters and hyperparameters are standard random variables in the Bayesian network of the model. Also, the rules for calculating probabilities according to this model are just the rules for calculating probabilities in this Bayesian network.

In what follows, we assume that the hyperparameters are fixed as $\Xise=\xise,\Xieae=\xieae,\Xie=\xie$. The following procedures (including both Bayesian and variational inference) can be generalised to also infer hyperparameters. However, our main reference \citep{friston_active_2015} and most publications on active inference also fix the hyperparameters.

\subsection{Bayesian Complete Posteriors}
\label{sec:binference}
\label{sec:plugin}
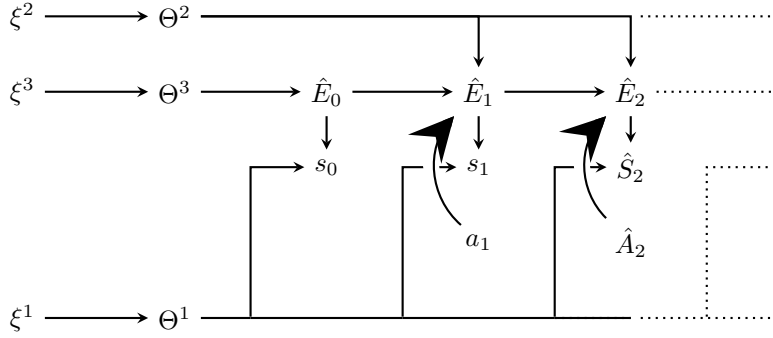
\begin{figure}%
\begin{center}
  \begin{tikzpicture}
    [->,>=stealth,auto,node distance=2cm,
    thick]
    \tikzset{
    hv/.style={to path={-| (\tikztotarget)}},
    vh/.style={to path={|- (\tikztotarget)}},
    }
    \tikzset{invi/.style={minimum width=0mm,inner sep=0mm,outer sep=0mm}}

    \node (e) [] {$\hE_1$};
    \node (e') [right of=e] {$\hE_{2}$};
    \node (s) [below of=e, node distance=1cm] {$s_1$};
    \node (s') [below of=e', node distance=1cm] {$\hS_{2}$};
    \node (a) [below of=s, node distance=1cm] {$a_1$};
    \node (a') [right of=a] {$\hA_{2}$};
    \node (m') [below of=a', node distance=1cm] {};
    \node (el) [left of=e] {$\hE_0$};
    \node (er) [right of=e'] {};
    \node (sl) [below of=el, node distance=1cm] {$s_0$};
    \node (sr) [below of=er, node distance=1cm] {};
    \node (al) [below of=sl, node distance=1cm] {};
    \node (ar) [right of=a'] {};
    \node (ml) [below of=al, node distance=1cm] {};
    \node (mr) [below of=ar, node distance=1cm] {};
    \node (th3) [left of=el] {$\Thetae$};
    \node (th2) [above of=th3, node distance=1cm] {$\Thetaeae$};
    \node (th1) [below of=th3, node distance=3cm] {$\Thetase$};

    \node (al3) [left of=th3] {$\xie$};
    \node (al2) [left of=th2] {$\xieae$};
    \node (al1) [left of=th1] {$\xise$};

    \node (th2') [above of=e', node distance=1cm] {};  
    \node (th2r) [right of=th2'] {};   
    
    \node (c0) [right of=th1, node distance=1cm,invi] {};
    \node (c1) [right of=c0,invi] {};
    \node (c2) [right of=c1,invi] {};
    \node (c3) [right of=c2,invi] {};
    \node (c3a) [right of=c2,node distance=1cm,invi] {};

    \path
      (al3) edge (th3)
      (al2) edge (th2)
      (al1) edge (th1)

      (th3) edge (el)
      (th2) edge[hv] (e)
      (th2) edge[hv] (e')
      (th2') edge[-,dotted] (th2r)
      (th1) edge[-] (c0)
      (c0) edge[vh] (sl)
      (c0) edge[-] (c1)
      (c1) edge[vh] (s)
      (c1) edge[-] (c2)
      (c2) edge[vh] (s')
      (c2) edge[-] (c3a)
      (c2) edge[-,dotted] (c3a)
      
      (c3) edge[-,dotted,vh] (sr)
      
      (e) edge node {} (e')
      (e) edge node {} (s)
      (e') edge node {} (s')
      (a) edge[bend left=45,line width=6pt,draw=white] node {} (e)   
      (a') edge[bend left=45,line width=6pt,draw=white] node {} (e')

      (a) edge[bend left=45] node {} (e)   
      (a') edge[bend left=45] node {} (e')

      (el) edge node {} (e)
      (el) edge node {} (sl)

      (e') edge[-,dotted] node {} (er)
      (m') edge[-,dotted] node {} (mr)

      ;
  \end{tikzpicture}
  \caption{Internal generative model with plugged in data up to $t=2$ with $\hS_0=s_0,\hS_1=s_1$ and $\hA_1=a_1$ as well as from now on fixed hyperparameters $\xi=(\xise,\xieae,\xie)$. 
Conditioning on the plugged in data leads to the posterior distribution $\q(\hs_\thT,\he_{0:\hT},\ha_\thT,\theta|sa_\pt,\xi)$.
  Predictions for future sensor values can be obtained by marginalising out other random variables e.g.\ to predict $\hS_2$ we would like to get $\q(\hs_2|s_0,s_1,a_1,\xi)$. Note however that this requires an assumption for the probability distribution over $\hA_2$.}
  \label{fig:genmodeldata}
\end{center}
\end{figure}

During action generation (i.e.\ within $\p(a|m)$) at time $t$, the agent has retained all its previously perceived sensor states and its previously performed actions in memory. The ``experience'' or data contained in its memory is thus $m_t=sa_\pt$. This data can be plugged into the generative model to obtain posterior probability distributions over all non-observed random variables. Also, the model can estimate the not yet observed sensor values $\hs_{t:\hT}$, past and future unobservable environment states $\he_{0:\hT}$, parameters $\theta$ and hyperparameters $\xi$. These estimations are done by setting:
\begin{equation}
  \hA_\tau = a_\tau, \text{for } \tau < t
\end{equation} 
and 
\begin{equation}
  \hS_\tau = s_\tau, \text{for } \tau < t.
\end{equation} 
as shown in \cref{fig:genmodeldata} for $t=2$. For these assignments to be generally possible, we need to choose $\shA$ and $\shS$ equal to $\sA$ and $\sS$ respectively. The resulting posterior probability distribution over all non-observed random variables is then, according to standard rules of calculating probabilities in a Bayesian network:
\begin{align}
\label{eq:posterior}
  \q(\hs_\thT,\he_{0:\hT},\ha_\thT,\theta|sa_\pt,\xi):&= \frac{\q(s_\pt,\hs_\thT,\he_{0:\hT},a_\pt,\ha_\thT,\theta,\xi)}{\int\sum_{\hs_\thT,\he_{0:\hT},\ha_\thT}\q(s_\pt,\hs_\thT,\he_{0:\hT},a_\pt,\ha_\thT,\theta,\xi)\diff \theta }.
\end{align}
Eventually, the agent needs to evaluate the consequences of its future actions. Just as it can update the model with respect to past actions and sensor values, the agent can update its evaluations with ``contemplated'' future action sequences $\ha_\thT$. For each such future action sequence $\ha_\thT$, the agent obtains a  distribution over the remaining random variables in the model: 
\begin{align}
\label{eq:posteriorgfa}
  \q(\hs_\thT,\he_{0:\hT},\theta|\ha_\thT,sa_\pt,\xi):&= \frac{\q(s_\pt,\hs_\thT,\he_{0:\hT},a_\pt,\ha_\thT,\theta,\xi)}{\int\sum_{\hs_\thT,\he_{0:\hT}}\q(s_\pt,\hs_\thT,\he_{0:\hT},a_\pt,\ha_\thT,\theta,\xi)\diff \theta }.
\end{align}
We call each such distribution a \textit{Bayesian complete posterior}. We choose the term complete posterior since the ``posterior'' by itself usually refers to the posterior distribution over the parameters and latent variables $\q(\theta,\he_\tm|sa_\pt,\xi)$ (we here call this a \textit{posterior factor}, see \cref{eq:posteriorgfa2}) and the posterior predictive distributions marginalise out the parameters and latent variables to get $\q(\hs_\thT|\ha_\thT,sa_\pt,\xi)$. The complete posteriors are probability distributions over all random variables in the generative model including parameters, latent variables, and future variables. In this sense the set of all (Bayesian) complete posteriors represents the complete knowledge state of the agent at time $t$ about consequences of future actions after updating the model with past actions and observed sensor values $sa_\pt$. At each time step the sequence of past actions and sensor values is extended from $sa_\pt$ to $sa_{\pt+1}$ (i.e.\ $m_t$ goes to $m_\tp$) and a new set of complete posteriors is obtained.  

All intrinsic motivations discussed in this article evaluate future actions based on quantities that can be derived from the corresponding complete posterior. 

It is important to note that the complete posterior can be factorised into a term containing the influence of past sensor values and actions (data). This factorisation can be made on the parameters $\theta$ and $\xi$, the environment states $\he_\pt$, predicted future environment states $\he_\thT$ and sensor values $\hs_\thT$ depending on the future actions $\ha_\thT$, and the estimated environment state $\he_\tm$ and $\theta$. Using the conditional independence 
\begin{align}
  SA_\pt &\ci \hS_\thT,\hE_\thT \mid\hA_\thT,\hE_\tm,\Theta,\Xi,
\end{align}
which can be identified (via $d$-separation \citep{pearl_causality_2000}) from the Bayesian network in \cref{fig:genmodeldata},
we can rewrite this as: 
\begin{align}
\label{eq:posteriorgfa2}
  \q(\hs_\thT,\he_{0:\hT},\theta|\ha_\thT,sa_\pt,\xi)&= \q(\hs_\thT,\he_\thT|\ha_\thT, \he_\tm,\theta)\q(\he_\pt, \theta|sa_\pt,\xi).
\end{align}
This equation represents the desired factorisation. This formulation separates complete posteriors into a predictive and a posterior factor. The predictive factor is given as part of the generative model (\cref{eq:genmodel})
\begin{align}
  \q(\hs_\thT,\he_\thT|\ha_\thT, \he_\tm,\theta)= \prod_{r=t}^\hT \q(\hs_r|\he_r,\thetase) \q(\he_r|\ha_r,\he_{r-1},\thetaeae)
\end{align}
and does not need to be updated through calculations at different time steps. This factor contains the dependence of the complete posterior on future actions. This dependency reflects that, under the given generative model, the consequences of actions for each combination of $\Theta$ and $\hE_\tm$ remain the same irrespective of experience. What changes when a new action and sensor value pair comes in is the distribution over the values of $\Theta$ and $\hE_\tm$ and with them the \emph{expectations} over consequences of actions. 

On the other hand, the posterior factor must be updated at every time step. In \cref{sec:postfactorBI}, we sketch the computation which shows that it involves a sum over $|\sE|^t$ elements. This calculation is intractable as time goes on and one of the reasons to use approximate inference methods like variational inference.    

Due to the above factorisation, we may only need to approximate the posterior factor $\q(\he_\pt, \theta|sa_\pt,\xi)$ and use the exact predictive factor if probabilities involving future sensor values or environment states are needed. 

This is
the approach taken e.g.\ in \citet{friston_active_2015}. However, it is also possible to directly approximate parts of the complete posterior involving random variables in both factors , e.g.\ by approximating $\q(\he_{0:\hT},\thetase|\ha_\thT,sa_\pt,\xi)$. This latter approach is taken
in \citet{friston_active_2016} and we see it again in \cref{eq:futurefactorizedpost} but in this publication the focus is on the former approach.

In the next section, we look at the special case of universal reinforcement learning before we go on to variational inference to approximate the posterior factor of the (Bayesian) complete posteriors.

\subsection{Connection to Universal Reinforcement Learning}
\label{sec:urlmodel}
In this section, we relate the generative model of \cref{eq:genmodel} and its posterior predictive distribution to those used by the Bayesian universal reinforcement learning agent. Originally, this agent is defined by \citet{hutter_universal_2005}. More recent work includes \citet{leike_nonparametric_2016} and (for the current purpose sufficient and particularly relevant) \citet{aslanides_universal_2017}.

Let us set $\shE=\sM$ with $\sM$ as in \cref{eq:memorystatespace} and let the agent identify each past $sa_\pt$ with a state of the environment, i.e.\:
\begin{align}
  \he_{t-1}=sa_\pt.
\end{align}
Under this definition the next environment state $\he_t$ is just the concatenation of the last environment state $sa_\pt$ with the next next action selected by the agent $\ha_t$ and the next sensor value $\hs_t$: 
\begin{align}
\he_t=\hs\ha_\pet=sa_\pt \hs\ha_t.
\end{align}
So given a next contemplated action $\bar{\ha}_t$ the next environment state $\he_t$ is already partially determined. What remains to be predicted is only the next sensor value $\hs_t$. 
Formally, this is reflected in the following derivation:
\begin{align}
  \q(\he_t|\bar{\ha}_t,\he_\tm,\thetaeae) :&= \q(\hs_t,\ha_t,\hs\ha_\pt|\bar{\ha}_t,sa_\pt,\thetaeae)\\
  &=\q(\hs_t|\ha_t,\hs\ha_\pt,\bar{\ha}_t,sa_\pt,\thetaeae) \q(\ha_t,\hs\ha_\pt|\bar{\ha}_t,sa_\pt,\thetaeae)\\
  &=\q(\hs_t|\ha_t,\hs\ha_\pt,\bar{\ha}_t,sa_\pt,\thetaeae) \delta_{\bar{\ha}_t}(\ha_t) \delta_{sa_\pt}(\hs\ha_\pt)\\
  &=\q(\hs_t|\bar{\ha}_t,sa_\pt,\thetaeae) \delta_{\bar{\ha}_t}(\ha_t) \delta_{sa_\pt}(\hs\ha_\pt).
\end{align}
This shows that in this case the model of the next environment state (the left hand side) is determined by the model of the next sensor value $\q(\hs_t|\bar{\ha}_t,sa_\pt,\thetaeae)$. 
So instead of carrying a distribution over possible models of the next environment state such an agent only needs to carry a distribution over models of the next sensor value. Furthermore, an additional model $\q(\hs|\he,\thetase)$ of the dependence of the sensor values on environment states parameterised by $\thetase$ is superfluous. The next predicted sensor value is already predicted by the model $\q(\hs_t|\ha_t,sa_\pt,\thetaeae)$. 
It is therefore possible to drop the parameter $\thetase$.

The parameter $\thetae$, for the initial environment state distribution, becomes a distribution over the initial sensor value since $\he_0=\hs_0$:
\begin{align}
  \q(\he_0|\thetae)=\q(\hs_0|\thetae).
\end{align}
We can then derive the posterior predictive distribution and show that it coincides with the one given in \citet{aslanides_universal_2017}.
For the complete posterior of \cref{eq:posteriorgfa2} we find:
\begin{align}
  \q(\hs_\thT,\he_{0:\hT},\theta|\ha_\thT,sa_\pt,\xi)&= 
  \q(\hs_\thT,\he_\thT|\ha_\thT, \he_\tm,\theta)\q(\he_\pt, \theta|sa_\pt,\xi) \tag{\ref{eq:posteriorgfa2} \text{ revisited}}\\
&= \q(\he_\thT|\hs_\thT,\ha_\thT, \he_\tm,\theta)\q(\hs_\thT|\ha_\thT, \he_\tm,\theta)\q(\he_\pt, \theta|sa_\pt,\xi)\\
  &= \q(\hs_\thT|\ha_\thT,sa_\pt,\theta) \q(\theta|sa_\pt,\xi) \prod_{\tau=0}^t \delta_{sa_\ptau}(\he_\tau) \prod_{\tau=t+1}^\hT \delta_{sa_\pt\hs\ha_{t:\tau}}(\he_\tau).
\end{align}
To translate this formulation into the notation of \citet{aslanides_universal_2017} first drop the representation of the environment state which is determined by the sensor values and actions anyway. This means that the complete posterior only needs to predict future sensor values and parameters. Formally, this means the complete posterior can be replaced without loss of generality: 
\begin{align}
  \q(\hs_\thT,\he_{0:\hT},\theta|\ha_\thT,sa_\pt,\xi) \rightarrow \q(\hs_\thT|\ha_\thT,sa_\pt,\theta) \q(\theta|sa_\pt,\xi).
\end{align}
To translate notations let $\theta \rightarrow \nu$; $\ha, a \rightarrow a$; $\hs,s \rightarrow e$. Also, set $\hT\rightarrow t$ because only one step futures are considered in universal reinforcement learning (this is due to the use of policies instead of future action sequences). Then, the equation for the posterior predictive distribution
\begin{align}
  \q(\hs_t|\ha_t, sa_\pt,\xi) = \int \q(\hs_t|\ha_t,sa_\pt,\theta) \q(\theta|sa_\pt,\xi) \diff \theta,
\end{align}
is equivalent to \citet[Eq. (5)]{aslanides_universal_2017} (the sum replaces the integral for a countable $\sTheta$):
\begin{align}
  \xi(e|ae_\pt,a) &= \sum_{\nu} p(e|\nu,ae_\pt,a) p(\nu|ae_\pt)\\
 \Leftrightarrow \xi(e) &= \sum_{\nu} p(e|\nu) p(\nu),
\end{align}
where we dropped the conditioning on $ae_\pt,a$ from the notation in the second line as done in the original (where this is claimed to improve clarity). Also note that $\xi(e)$ would be written $\q(e|\xi)$ in our notation. In the universal reinforcement learning literature parameters like $\theta$ (or $\nu$) and $\xi$ are sometimes directly used to denote the probability distribution that they parameterise. 

Updating of the posterior $\q(\theta|sa_\pt,\xi)$ in response to new data also coincides with updating of the weights $p(\nu)$: 
\begin{align}
  \q(\theta|sa_\pet,\xi) &= \frac{\q(\theta,s_t|a_t,sa_\pt,\xi)}{\q(s_t|a_t,sa_\pt,\xi)}\\
  &= \frac{\q(s_t|a_t,sa_\pt,\theta,\xi) \q(\theta|a_t,sa_\pt,\xi)}{\q(s_t|a_t,sa_\pt,\xi)}\\
  &=\frac{\q(s_t|a_t,sa_\pt,\theta) \q(\theta|sa_\pt,\xi)}{\q(s_t|a_t,sa_\pt,\xi)}\\
  &=\frac{\q(s_t|a_t,sa_\pt,\theta)}{\q(s_t|a_t,sa_\pt,\xi)} \q(\theta|sa_\pt,\xi). \label{eq:urlupdate}
\end{align}
The first two lines are general. From the second to third we used
\begin{align}
  S_t \ci \Xi | A_t,SA_\pt,\Theta
\end{align}
and
\begin{align}
  \Theta \ci A_t |SA_\pt, \Xi
\end{align}
which follow from the Bayesian network structure \cref{fig:genmodel}. In the notation of \citet{aslanides_universal_2017} \cref{eq:urlupdate} becomes
\begin{align}
  p(\nu|e) = \frac{p(e|\nu)}{p(e)} p(\nu).
\end{align}
This shows that assuming the same model class $\sTheta$ the predictions and belief updates of an agent using the Bayesian complete posterior of \cref{sec:binference} are the same as those of the Bayesian universal reinforcement learning agent. Action selection can then be performed just as in \citet{aslanides_universal_2017} as well. This is done by selecting policies. In the present publication we instead select action sequences directly. However, in both cases the choice maximises the value predicted by the model. More on this in \cref{sec:actionselection}.

\subsection{Approximate Complete Posteriors}
\label{sec:approxpostandvi}
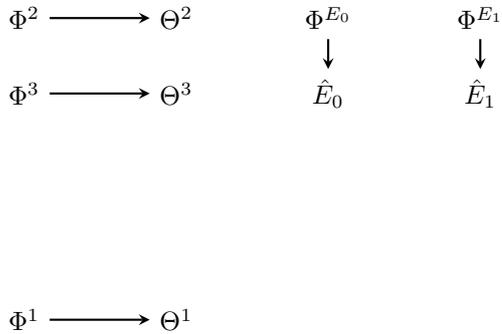
\begin{figure}%
\begin{center}
  \begin{tikzpicture}
    [->,>=stealth,auto,node distance=2cm,
    thick]
    \tikzset{
    hv/.style={to path={-| (\tikztotarget)}},
    vh/.style={to path={|- (\tikztotarget)}},
    }
    \tikzset{invi/.style={minimum width=0mm,inner sep=0mm,outer sep=0mm}}

    \node (e) [] {$\hE_1$};
    \node (s) [below of=e, node distance=1cm] {};
    \node (s') [below of=e', node distance=1cm] {};
    \node (a) [below of=s, node distance=1cm] {};
    \node (a') [right of=a] {};
    \node (m') [below of=a', node distance=1cm] {};
    \node (el) [left of=e] {$\hE_0$};
    \node (er) [right of=e'] {};
    \node (sl) [below of=el, node distance=1cm] {};
    \node (sr) [below of=er, node distance=1cm] {};
    \node (al) [below of=sl, node distance=1cm] {};
    \node (ar) [right of=a'] {};
    \node (ml) [below of=al, node distance=1cm] {};
    \node (mr) [below of=ar, node distance=1cm] {};
    \node (th3) [left of=el] {$\Thetae$};
    \node (th2) [above of=th3, node distance=1cm] {$\Thetaeae$};
    \node (th1) [below of=th3, node distance=3cm] {$\Thetase$};

    \node (al3) [left of=th3] {$\Phie$};
    \node (al2) [left of=th2] {$\Phieae$};
    \node (al1) [left of=th1] {$\Phise$};

    \node (phie0) [above of=el,node distance=1cm] {$\Phi^{E_0}$};
    \node (phie1) [above of=e,node distance=1cm] {$\Phi^{E_1}$};
    \node (th2r) [right of=th2'] {};   
    
    \node (c0) [right of=th1, node distance=1cm,invi] {};
    \node (c1) [right of=c0,invi] {};
    \node (c2) [right of=c1,invi] {};
    \node (c3) [right of=c2,invi] {};
    \node (c3a) [right of=c2,node distance=1cm,invi] {};

    \path
      (al3) edge (th3)
      (al2) edge (th2)
      (al1) edge (th1)

      (phie0) edge (el)
      (phie1) edge (e)

      ;
  \end{tikzpicture}
  \caption{Bayesian network of the approximate posterior factor at $t=2$. The variational parameters $\Phise,\Phieae,\Phie$ and $\Phipt=(\Phinull,\Phione)$ are positioned so as to indicate what dependencies and nodes they replace in the generative model in \cref{fig:genmodel}.} 
  \label{fig:recmodel}
\end{center}
\end{figure}

\begin{figure}%
\begin{center}
  \begin{tikzpicture}
    [->,>=stealth,auto,node distance=2cm,
    thick]
    \tikzset{
    hv/.style={to path={-| (\tikztotarget)}},
    vh/.style={to path={|- (\tikztotarget)}},
    }
    \tikzset{invi/.style={minimum width=0mm,inner sep=0mm,outer sep=0mm}}

    \node (e) [] {$\hE_1$};
    \node (e') [right of=e] {$\hE_{2}$};
    \node (s) [below of=e, node distance=1cm] {};
    \node (s') [below of=e', node distance=1cm] {$\hS_2$};
    \node (a) [below of=s, node distance=1cm] {};
    \node (a') [right of=a] {$\ha_2$};
    \node (m') [below of=a', node distance=1cm] {};
    \node (el) [left of=e] {$\hE_0$};
    \node (er) [right of=e'] {};
    \node (sl) [below of=el, node distance=1cm] {};
    \node (sr) [below of=er, node distance=1cm] {};
    \node (al) [below of=sl, node distance=1cm] {};
    \node (ar) [right of=a'] {};
    \node (ml) [below of=al, node distance=1cm] {};
    \node (mr) [below of=ar, node distance=1cm] {};
    \node (th3) [left of=el] {$\Thetae$};
    \node (th2) [above of=th3, node distance=1cm] {$\Thetaeae$};
    \node (th1) [below of=th3, node distance=3cm] {$\Thetase$};

    \node (th2u) [above of=th2,node distance=.5cm,invi] {};
    
    \node (al3) [left of=th3] {$\Phie$};
    \node (al2) [left of=th2] {$\Phieae$};
    \node (al1) [left of=th1] {$\Phise$};

    \node (phie0) [above of=el,node distance=1cm] {$\Phi^{E_0}$};
    \node (phie1) [above of=e,node distance=1cm] {$\Phi^{E_1}$};
    \node (th2r) [right of=th2'] {};   
    
    \node (c0) [right of=th1, node distance=1cm,invi] {};
    \node (c1) [right of=c0,invi] {};
    \node (c2) [right of=c1,invi] {};
    \node (c3) [right of=c2,invi] {};
    \node (c3a) [right of=c2,node distance=1cm,invi] {};

    \path
      (al3) edge (th3)
      (al2) edge (th2)
      (al1) edge (th1)

      (phie0) edge (el)
      (phie1) edge (e)
      (th1) edge[-] (c0)
      (c0) edge[-] (c1)
      (c1) edge[-] (c2)
      (c2) edge[vh] (s')
      
      (th2) edge[-] (th2u)
      (th2u) edge[hv] (e')
      (th2u) edge[-,dotted,hv] (er)
      (c2) edge[-] (c3a)
      (c2) edge[-,dotted] (c3)

      (c3) edge[-,dotted,vh] (sr)
      
      (e) edge node {} (e')
      (e') edge node {} (s')
      (a') edge[bend left=45,line width=6pt,draw=white] node {} (e')

      (a') edge[bend left=45] node {} (e')

      (e') edge[-,dotted] node {} (er)

      ;
  \end{tikzpicture}
  \caption{Bayesian network of the approximate complete posterior of \cref{eq:apposteriorxi} at $t=2$ for the future actions $\ha_\thT$. Only $\hE_\tm, \Thetase,\Thetaeae$ and the future action $\ha_\thT$ appear in the predictive factor and influence future variables. In general there is one approximate complete posterior for each possible sequence $\ha_\thT$ of future actions.}
  \label{fig:recpred}
\end{center}
\end{figure}
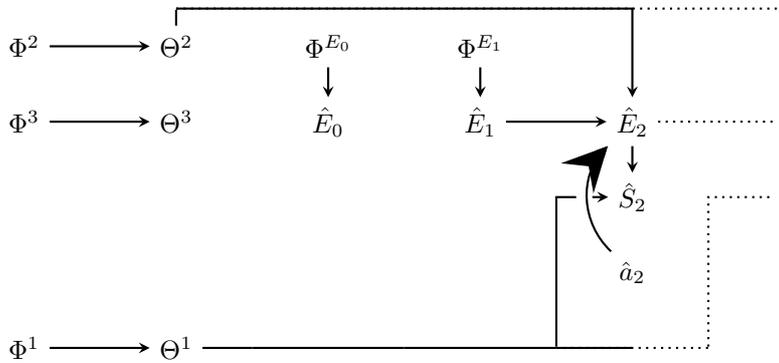

As mentioned in the last section, the complete posterior can be approximated via variational inference \citep[see ][]{attias_variational_1999,winn_variational_2005,bishop_pattern_2011,blei_variational_2017}. There are alternative methods such as belief propagation, expectation propagation \citep{minka_expectation_2001,vehtari_expectation_2014}, and sampling-based methods \citep{lunn_winbugs_2000,bishop_pattern_2011}, but active inference commits to variational inference by framing inference as variational free energy minimisation \citep{friston_active_2015}. Variational free energy (\cref{eq:fesimple}) is just the negative evidence lower bound (ELBO) of standard variational inference \citep[e.g.][]{blei_variational_2017}. In the following, we show how the complete posterior can be approximated via variational inference.

The idea behind variational inference is to use a simple family of probability distributions and identify the member of that family which approximates the true complete posterior best. This turns inference into an optimisation problem. According to \citet{wainwright_graphical_2007} this reformulation as an optimisation problem is the essence of variational methods. If the family of distributions is chosen such that it includes the complete posterior then the optimisation will eventually lead to the same result as Bayesian inference. However, one advantage of the formulation as an optimisation is that it can also be performed over a family of probability distributions that is simpler than the family that includes the actual complete posterior. This is what turns variational inference into an approximate inference procedure. Usually, the (simpler) families of probability distributions are chosen as products of independent distributions. 

Recalling \cref{eq:posteriorgfa2}, the complete posterior as a product of a predictive and a posterior factor is:
\begin{align*}
  \q(\hs_\thT,\he_{0:\hT},\theta|\ha_\thT,sa_\pt,\xi)&= \q(\hs_\thT,\he_\thT|\ha_\thT, \he_\tm,\theta)\q(\he_\pt, \theta|sa_\pt,\xi). \tag{\ref{eq:posteriorgfa2} revisited}
\end{align*}
This product is the main object of interest. We want to approximate the formula with a probability distribution that lets us (tractably) calculate the posteriors required by a given intrinsic motivation, which can consequently be used for action selection. %

As mentioned before, to approximate the complete posterior we here approximate only the posterior factor and use the given generative model's predictive factor as is done in \citet{friston_active_2015}\footnote{A close inspection of \citet[Eq. (9)]{friston_active_2015} shows that the approximate complete posterior that ends up being evaluated by the action-value function is the one we discuss in \cref{eq:apposteriorxi}. It uses the predictive factor to get the probabilities $\r(\he_\thT|\ha_\thT,\he_\tm,\phi)$ of future environment states. However, the approximate posterior in \citet[Eq.(10)]{friston_active_2015} uses a factorisation of all future environment states like the one we give in \cref{eq:futurefactorizedpost}. The probabilities of future environment states in that posterior are not used anywhere in \citet{friston_active_2015}. In principle, they could be used as is done in \citet[Eq. (2.6)]{friston_active_2016} where the complete posterior of \cref{eq:futurefactorizedpost} is used in the action-value function. Both approaches are possible.}  
The approximate posterior factor is then combined with the exact predictive factor to get the approximate complete posterior. Let us write $\r(\he_\pt,\theta|\phi)$ for the approximate posterior factor (\cref{fig:recmodel}), defined as:
\begin{align}
\label{eq:justappost}
  \r(\he_\pt,\theta|\phi):&=\r(\he_\pt|\phipt)\r(\theta|\phi)\\
  :&=\prod_{\tau=0}^{t-1} \r(\he_\tau|\phitau)  \prod_{i=1}^3 \r(\theta^i|\phi^i).
\end{align}
As we can see it models each of the random variables that the posterior factor ranges over as independent of all others. This is called a \textit{mean field} approximation.
Then, the approximate complete posterior (\cref{fig:recpred}) is:
\begin{align}
\label{eq:apposteriorxi}
  \r(\hs_\thT,\he_{0:\hT},\theta|\ha_\thT,\phi):&= \q(\hs_\thT,\he_\thT|\ha_\thT, \he_\tm,\theta)\r(\he_\pt, \theta|\phi).
\end{align}
Note that the variational parameter absorbs the hyperparameter $\xi$ as well as the past sensor values and actions $sa_\pt$. The parameter does not absorb future actions which are part of the predictive factor. The dependence on future actions needs to be kept if we want to select actions using the approximate complete posterior. 

We have: 
\begin{align}
  \r(\hs_\thT,\he_{0:\hT},\theta|\ha_\thT,\phi) \approx \q(\hs_\thT,\he_{0:\hT},\theta|\ha_\thT,sa_\pt,\xi)
\end{align}
if 
\begin{equation}
  \r(\he_\pt, \theta|\phi) \approx \q(\he_\pt, \theta|sa_\pt,\xi).
\end{equation} 
This approximation can be achieved by standard variational inference methods. 

For those interested more in the approximation of the complete posterior as in \citet{friston_active_2016}, we provide the used family of factorised distributions. It must be noted that the agent in this case carries a separate approximate posterior for each possible complete action sequence $\ha_\allt$. For predictions of environment states, it does not use the predictive factor, but instead looks at the set of generative models compatible with the past. For each of those, the agent considers all environment states at different times as independent. The approximate posteriors, compatible with a past sequence of actions $a_\pt$, are of the form:
\begin{align}
\label{eq:futurefactorizedpost}
  \r(\hs_\thT,\he_{0:\hT},\thetase|\ha_\thT,a_\pt,\phise) = \q(\hs_\thT|\he_\thT,\thetase)\prod_{\tau=0}^\hT \r(\he_\tau|\ha_\thT,a_\pt,\phitau) \r(\thetase|\phise).
\end{align}
Note also that the relation between sensor values and environment states is still provided by the generative models' sensor dynamics $\q(\hs_\thT|\he_\thT,\thetase)$. In this article however, we focus on the approach in \citet{friston_active_2015} which requires only one approximate posterior at time $t$ since future actions only occur in the predictive factors which we do not approximate.

We define the relative entropy (or $\KL$-divergence) between the approximate and the true posterior factor:
\begin{align}
  \KL[\r(\hE_\pt,\Theta|\phi)||\q(\hE_\pt,\Theta|sa_\pt,\xi)]:= \sum_{\he_\pt}\int \r(\he_\pt,\theta|\phi) \log \frac{\r(\he_\pt,\theta|\phi)}{\q(\he_\pt,\theta|sa_\pt,\xi)} \diff \theta.
\end{align}
Note that, we indicate the variables that are summed over by capitalising them. 
The $\KL$-divergence quantifies the difference between the two distributions. It is non-negative, and only zero if the approximate and the true posterior factor are equal \citep[see e.g.~][]{cover_elements_2006}.

The variational free energy, also known as the (negative) evidence lower bound (ELBO) in variational inference literature, is defined as:
\begin{align}
\label{eq:fesimple}
  \F[\xi,\phi,sa_\pt]:&=\sum_{\he_\pt} \int \r(\he_\pt,\theta|\phi) \log \frac{\r(\he_\pt,\theta|\phi)}{\q(s_\pet,\he_\pt,\theta|a_\pt,\xi)} \diff \theta\\
  &= - \log \q(s_\pt|a_\pt,\xi) +  \KL[\r(\hE_\pt,\Theta|\phi)||\q(\hE_\pt,\Theta|sa_\pt,\xi)] \label{eq:elbo2}
\end{align}
The first term in \cref{eq:elbo2} is the surprise of negative log evidence. For a fixed hyperparameter $\xi$ it is a constant. Minimising the variational free energy therefore directly minimises the $\KL$-divergence between the true and the approximate posterior factor given $sa_\pt$ and $\xi$. 

In our case, variational inference amounts to solve the optimisation problem:
\begin{align}
\label{eq:vi}
  \phis_{sa_\pt,\xi}:=\argmin_\phi \F[\phi,sa_\pt,\xi].
\end{align}
This optimisation is a standard problem. See \citet{bishop_pattern_2011,blei_variational_2017} for ways to solve it.

The resulting variational parameters $\phis_{sa_\pt,\xi}=(\phinull_{sa_\pt,\xi},...,\phitm_{sa_\pt,\xi},\phise_{sa_\pt,\xi},\phieae_{sa_\pt,\xi},\phie_{sa_\pt,\xi})$ define the approximate posterior factor. The variational parameters, together with the exact predictive factors, allow us to compute the approximate complete posteriors for each sequence of future actions $\ha_\thT$:
\begin{align}
  \r(\hs_\thT,\he_{0:\hT},\theta|\ha_\thT,\phis_{sa_\pt,\xi}) &= \q(\hs_\thT,\he_\thT|\ha_\thT, \he_\tm,\theta)\r(\he_\pt, \theta|\phis_{sa_\pt,\xi})\\
  &\approx \q(\hs_\thT,\he_{0:\hT},\theta|\ha_\thT,sa_\pt,\xi).
\end{align}

In the next section, we look at action selection as the second component of action generation. To this end, we show how to evaluate sequences of future actions $\ha_\thT$ by evaluating either Bayesian complete posteriors or the approximate complete posteriors.

\section{Action Selection Based on Intrinsic Motivations}
\label{sec:aselectandim}

\subsection{Intrinsic Motivation and Action-Value Functions}

The previous section resulted in sets of Bayesian or approximate complete posteriors. Independently of whether a complete posterior is the approximate or the  Bayesian version, it represents the entire knowledge of the agent about the consequences of the sequence of future actions $\ha_\thT$ that is associated with it.
In order to evaluate sequences of future actions the agent can only rely on its knowledge which suggests that all such evaluations should depend solely on complete posteriors. One could argue that the motivation might also depend directly on the memory state containing $sa_\pt$. We here take a position somewhat similar to the one proposed by \citet{schmidhuber_formal_2010} that intrinsic motivations concerns the ``learning of a better world model''. We consider the complete posterior as the current world model and assume that intrinsic motivations depend only on this model and not on the exact values of past sensor values and actions. As we will see this assumption is also enough to capture the three intrinsic motivations that we discuss here. This level of generality is sufficient for our purpose of extending the free energy principle. Whether it sufficient for a final and general intrinsic motivation definition is beyond the scope of this publication.

Complete posteriors are essentially conditional probability distributions over $\shS^{\hT-t+1}\times \shE^{\hT+1}\times \sTheta$ given elements of $\shA^{\hT-t+1}$. A necessary (but not sufficient) requirement for intrinsic motivations in our context (agents with generative models) is then that they are functions on the space of such conditional probability distributions. Let $\Delta_{\shS^{\hT-t+1}\times \shE^{\hT+1}\times \sTheta|\shA^{\hT-t+1}}$ be the space of conditional probability distributions over $\shS^{\hT-t+1}\times \shE^{\hT+1}\times \sTheta$ given elements of $\shA^{\hT-t+1}$. Then an \textit{intrinsic motivation} is a function $\mot: \Delta_{\shS^{\hT-t+1}\times \shE^{\hT+1}\times \sTheta|\shA^{\hT-t+1}} \times \shA^{\hT-t+1} \rightarrow \mathbb{R}$ taking a probability distribution $\d(.,.,.|.) \in \Delta_{\shS^{\hT-t+1}\times \shE^{\hT+1}\times \sTheta|\shA^{\hT-t+1}}$ and a given future actions sequence $\ha_\thT \in \shA^{\hT-t+1}$ to a real value $\mot(\d(.,.,.|.),\ha_\thT) \in \mathbb{R}$. 
We can then see that the Bayesian complete posterior $\q(\hs_\thT,\he_\allht,\theta|\ha_\thT,sa_\pt,\xi)$ for a fixed past $sa_\pt$ written as $\q(.,.,.|.,sa_\pt,\xi)$ provides such conditional probability distribution. Similarly, every member of the family of distributions used to approximate the Bayesian complete posterior via variational inference $\r(\hs_\thT,\he_{0:\hT},\theta|\ha_\thT,\phi)$ written as $\r(.,.,.|.,\phi)$ also provides such a conditional probability distribution. It will become important when discussing active inference that the optimised value $\phis_{sa_\pt,\xi}$ of the variational parameters as well as any other value of the variational parameters $\phi$ define an element with the right structure to be evaluated together with a set of future actions by an intrinsic motivation function.

Using intrinsic motivation functions we then define two kinds of induced action-value functions. These are similar to value functions in reinforcement learning.
\footnote{The main difference is that the action-value functions here evaluate sequences of future actions as opposed to policies. This is the prevalent practice in active inference literature including \citet{friston_active_2015} and we therefore follow it here. 
} 
The first is the \textit{Bayesian action-value function} (or functional):
\begin{equation}
\label{eq:biactionvalue}
  \hQ(\ha_\thT,sa_\pt,\xi):= \mot(\q(.,.,.|.,sa_\pt,\xi),\ha_\thT).
\end{equation}
In words the Bayesian action-value function $\hQ(\ha_\thT,sa_\pt,\xi)$ infers the set of Bayesian complete posteriors of past experience $sa_\pt$ and then evaluates the sequence of future actions $\ha_\thT$ according to the intrinsic motivation function $\mot$. 

The \textit{variational action-value function} is defined as\footnote{We abuse notation here by reusing the same symbol $\hQ$ for the variational action-value function as for the Bayesian action-value function. However, in this publication the argument ($sa_\pt,\xi$ or $\phi$) always indicates which one is meant.}:
\begin{equation}
  \hQ(\ha_\thT,\phi):= \mot(\r(.,.,.|.,\phi),\ha_\thT).
\end{equation}
So the variational action-value function $\hQ(\ha_\thT,\phi)$ directly takes the conditional probability distribution defined by variational parameter $\phi$ and evaluates the sequence of future actions $\ha_\thT$ according to $\mot$. Unlike in the Bayesian case no inference takes place during the evaluation of $\hQ(\ha_\thT,\phi)$. 

At the same time, after variational inference, if we plug in $\phis_{sa_\pt,\xi}$ for $\phi$ we have:
\begin{align}
\label{eq:actionvalueapprox}
  \hQ(\ha_\thTa,\phis_{sa_\pt,\xi})\approx \hQ(\ha_\thTa,sa_\pt,\xi).
\end{align}

Note that the reason we have placed a hat on $\hQ$ is that, even in the Bayesian case, it is usually not the optimal action-value function but instead is an estimate based on the current knowledge state represented by the complete posteriors of the agent.

Also note that some intrinsic motivations (e.g.\ empowerment) evaluate e.g.\ the next $n$ actions by using predictions reaching $n+m$ steps into the future. This means that they need all complete posteriors for $\ha_{t:t+n+m-1}$ but only evaluate the actions $\ha_{t:t+n-1}$. In other words they cannot evaluate actions up to their generative model's time-horizon $\hT$ but only until a shorter time-horizon $\hTa=\hT-m$ for some natural number $m$. When necessary we indicate such a situation by only passing shorter future action sequences $\ha_\thTa$ to the action-value function, in turn, the intrinsic motivation function. The respective posteriors keep the original time horizon $\hT > \hTa$.

\subsection{Deterministic and Stochastic Action Selection}
\label{sec:actionselection}
We can then select actions simply by picking the first action in the sequence $\ha_\thT$ that maximises the Bayesian action-value function:
\begin{align}
\label{eq:argmaxaction}
  \ha_\thT^*(m_t):=\ha_\thT^*(sa_\pt):=\argmax_{\ha_\thT} \hQ(\ha_\thT,sa_\pt,\xi)
\end{align}
and set
\begin{align}
  \ha^*(m_t):=\ha_t^*(m_t).
\end{align}
or for the variational action value function:
\begin{align}
  \ha_\thT^*(m_t):=\ha_\thT^*(\phis_{sa_\pt,\xi}):=\argmax_{\ha_\thT} \hQ(\ha_\thT,\phis_{sa_\pt,\xi}).
\end{align}
and set
\begin{align}
  \ha^*(m_t):=\ha_t^*(m_t).
\end{align}

This then results in a deterministic action generation $\p(a|m)$:
\begin{align*}
  \p(a_t|m_t):=\delta_{\ha^*(m_t)}(a_t).
\end{align*}

We note here that in the case of universal reinforcement learning the role of $\hQ(\ha_\thT,sa_\pt,\xi)$ is played by $V^\pi_\xi(sa_\pt)$. There $\pi$ is a policy that selects actions in dependence on the entire past $sa_\pt$ and $\xi$ parameterises the posterior just like in the present publication. The $\argmax$ in \cref{eq:argmaxaction} selects a policy instead of an action sequence and that policy is used for the action generation.

A possible stochastic action selection that is important for active inference is choosing the action according to a so called softmax policy \citep{sutton_reinforcement_1998}:
\begin{align}
\label{eq:softmax}
  \p(a_t|m_t):=\sum_{\ha_{\tp:\hT}}\frac{1}{Z(\gamma,sa_\pt,\xi)} e^{\gamma \hQ(\ha_\thT,sa_\pt,\xi)}
\end{align}
where:
\begin{align}
  Z(\gamma,sa_\pt,\xi):= \sum_{\ha_\thT} e^{\gamma \hQ(\ha_\thT,sa_\pt,\xi)}
\end{align}
is a normalisation factor. Note that we are marginalising out later actions in the sequence $\ha_\thT$ to get a distribution only over the action $\ha_t$.
For the variational action-value function this becomes:
\begin{align}
\label{eq:softmax2}
  \p(a_t|m_t):=\sum_{\ha_{\tp:\hT}}\frac{1}{Z(\gamma,\phis_{sa_\pt,\xi})} e^{\gamma \hQ(\ha_\thT,\phis_{sa_\pt,\xi})}
\end{align}
where:
\begin{align}
  Z(\gamma,\phis_{sa_\pt,\xi}):= \sum_{\ha_\thT} e^{\gamma \hQ(\ha_\thT,\phis_{sa_\pt,\xi})}.
\end{align}
Since it is relevant for active inference (see \cref{sec:activeinference}), note that the softmax distribution over future actions can also be defined for arbitrary $\phi$ and not only for the optimised $\phis_{sa_\pt,\xi}$. At the same time, the softmax distribution for the optimised $\phi_{sa_\pt,\xi}$ clearly also approximates the softmax distribution of the Bayesian action-value function.

Softmax policies assign action sequences with higher values of $\hQ$ higher probabilities. They are often used as a replacement for the deterministic action selection to introduce some exploration. Here, lower $\gamma$ leads to higher exploration; conversely, in the limit where $\gamma \rightarrow \infty$ the softmax turns into the deterministic action selection. From an intrinsic motivation point of view such additional exploration should be superfluous in many cases since many intrinsic motivations try to directly drive exploration by themselves. Another interpretation of such a choice is to see $\gamma$ as a trade-off factor between the processing cost of choosing an action precisely and achieving a high action-value. The lower $\gamma$, the higher the cost of precision. This leads to the agent more often taking actions that do not attain maximum action-value. 

We note that the softmax policy is not the only possible stochastic action selection mechanism. Another option discussed in the literature is Thompson sampling \citep{ortega_minimum_2010,ortega_generalized_2014,aslanides_universal_2017}. In our framework this corresponds to a two step action selection procedure where we first sample an environment and parameter pair $(\bar{\he}_\tm,\bar{\theta})$ from a posterior factor (Bayesian or variational) 
\begin{align}
  (\bar{\he}_\tm,\bar{\theta}) \sim \d(\hE_\tm,\Theta|sa_\pt,\xi)
\end{align}
then plug the according predictive factor $\q(\hs_\thT,\he_\thT|\ha_\thT,\bar{\he}_\tm,\bar{\theta})$ into the action value function
\begin{equation}
  \hQ(\ha_\thT,sa_\pt,\xi):=\mot(\q(.,.|.,\bar{\he}_\tm,\bar{\theta}),\ha_\thT).
\end{equation} 
This allows intrinsic motivations that only evaluate the probability distribution over future sensor values $\hS_\thT$ and environment states $\hE_\thT$. However, it rules out those that evaluate the posterior probability of environment parameters $\Theta$ because we sample a specific $\bar{\theta}$.

\subsection{Intrinsic Motivations}
Now, we look at some intrinsic motivations including the intrinsic motivation part underlying Friston's active inference. 

In the definitions, we use  $\d(.,.,.|.) \in \Delta_{\shS^{\hT-t+1}\times \shE^{\hT+1}\times \sTheta|\shA^{\hT-t+1}}$ as a generic conditional probability distribution. The generic symbol $\d$ is used since it represents both Bayesian complete posteriors and approximate complete posteriors. In fact, the definitions of the intrinsic motivations are agnostic with respect to the method used to obtain a complete posterior. In the present context, it is important that these definitions are general enough to induce both Bayesian and variational action-value functions. We usually state the definition of the motivation function using general expressions (e.g.\ marginalisations) derived from $\d(,.,.|.)$. Also, we look at how they can be obtained from Bayesian complete posteriors to give to the reader an intuition for the computations involved in applications. The approximate complete posterior usually makes these calculations easier and we will present an example of this. 

\subsubsection{Free Energy Principle}
\label{sec:fep}
Here, we present the non-variational Bayesian inference versions for the expressions that occur in the ``expected free energy'' in \citet{friston_active_2015,friston_active_curiosity_2017}. These papers only include approximate expressions after variational inference. Most of the expressions we give here can be found in \citet{friston_graphical_2017}. The exception is \cref{eq:infogain}, which can be obtained from an approximate term in \citet{friston_active_curiosity_2017} in the same way that the non-variational Bayesian inference terms in \citet{friston_graphical_2017} are obtained from the approximate ones in \citet{friston_active_2015}.

In the following, we can set $\hTa=\hT$, since actions are only evaluated with respect to their immediate effects.

According to \citet[Eq. (A.2) supplementary material]{friston_graphical_2017}, the ``expected free energy'' is just the future conditional entropy of sensor values\footnote{The original text refers to this as the ``expected entropy of outcomes'', not the expected conditional entropy of outcomes. Nonetheless, the associated Equation (A.2) in the original is identical to ours.} given environment states. Formally, this is (with a negative sign to make minimising expected free energy equivalent to maximising the action-value function):
\begin{align}
\label{eq:fepentropy}
\mot(\d(.,.,.|.),\ha_\thT) :&=  \sum_{\he_\thT} \d(\he_\thT|\ha_\thT) \sum_{\hs_\thT} \d(\hs_\thT|\he_\thT) \log \d(\hs_\thT|\he_\thT)\\
&= - \sum_{\he_\thT} \d(\he_\thT|\ha_\thT) \HS_{\d}(\hS_\thT|\he_\thT)\\
&=-\HS_{\d}(\hS_\thT|\hE_\thT,\ha_\thT).
\end{align}
Note that, we indicate the probability distribution $\d$ used to calculate entropies $\HS_{\d}(X)$ or mutual informations $\I_{\d}(X:Y)$ in the subscript. Furthermore,we indicate the variables that are summed over with capital letters and those that are fixed (e.g.\ $\ha_\thT$ above) with small capital letters. 

In the case where $\d(.,.,.|.)$ is the Bayesian complete posterior $\q(.,.,.|.,sa_\pt,\xi)$, it uses the predictive distribution of environment states $\q(\he_\thT|\ha_\thT,sa_\pt,\xi)$ and the posterior of the conditional distribution of sensor values given environment states $\q(\hs_\thT|\he_\thT,sa_\pt,\xi)$. As we see next, both distributions can be obtained from the Bayesian complete posterior. 

The former distribution is a familiar expression in hierarchical Bayesian models and corresponds to a posterior predictive distribution or predictive density \citep[cmp. e.g.][Eq.(3.74)]{bishop_pattern_2011} that can be calculated via:
\begin{align}
  \q(\he_\thT|\ha_\thT,sa_\pt,\xi)&=   
  \int \sum_{\hs_\thT,\he_\pt} \q(\hs_\thT,\he_{0:\hT},\theta|\ha_\thT,sa_\pt,\xi) \diff \theta \\
&=  \int \sum_{\hs_\thT,\he_\pt} \q(\hs_\thT,\he_\thT|\ha_\thT, \he_\tm,\theta)\q(\he_\pt, \theta|sa_\pt,\xi) \diff \theta \\
&=  \int \sum_{\he_\tm} \q(\he_\thT|\ha_\thT, \he_\tm,\theta) \q(\he_\tm, \theta|sa_\pt,\xi) \diff \theta,
\label{eq:postprede}
\end{align}
where we split the complete posterior into the predictive and posterior factor and then marginalised out environment states $\he_{\pt-1}$ since the predictive factor does not depend on them. Note that in practice, this marginalisation corresponds to a sum over $|\sE|^{t-1}$ terms and therefore has a computational cost that grows exponential in time.  
However, if we use the approximate complete posterior such that $\d(.,.,.|.)=\r(.,.,.|.,\phi)$, we see from \cref{eq:apposteriorxi}, that $\q(\he_\pt, \theta|sa_\pt,\xi)$ is replaced by $\r(\he_\pt, \theta|\phi)$ which is defined as (\cref{eq:justappost}):
\begin{align}
\r(\he_\pt, \theta|\phi) :=\prod_{\tau=0}^{t-1} \r(\he_\tau|\phitau)  \prod_{i=1}^3 \r(\theta^i|\phi^i).
\end{align}
This means that $\r(\he_\tm, \theta|\phi)$ is just $\r(\he_\tm|\phitm)\r(\theta|\phi)$, which we obtain directly from the variational inference without any marginalisation. If Bayesian inference increases in computational cost exponentially in time, this simplification leads to a significant advantage.
This formulation leaves an integral over $\theta$ or, more precisely, a triple integral over the three $\thetase,\thetaeae,\thetae$. However, if the $\q(\theta^i|\xi^i)$ are chosen as conjugate priors to $\q(\hs|\he,\thetase),\q(\he'|\ha',\he,\thetaeae),\q(\he_0|\thetae)$ respectively, then these integrals can be calculated analytically (compare the similar calculation of $\q(\he_\pt, \theta|sa_\pt,\xi)$ in \cref{sec:postfactorBI}). The remaining computational problem is only the sum over all $\he_{t-1}$. 

The latter term (the posterior conditional distribution over sensor values given environment states) can be obtained via 
\begin{align}
  \q(\hs_\thT|\he_\thT,sa_\pt,\xi)&= \q(\hs_\thT|\he_\thT,\ha_\thT,sa_\pt,\xi)\\
  &=\frac{\q(\hs_\thT,\he_\thT|\ha_\thT,sa_\pt,\xi)}{\q(\he_\thT|\ha_\thT,sa_\pt,\xi)}.
\label{eq:postsge}
\end{align}
Here, the first equation holds since 
\begin{equation}
  \hS_\thT \ci \hA_\thT \mid \hE_\thT,SA_\pt.
\end{equation} 
Both numerator and denominator can be obtained from the complete posterior via marginalisation as for the former term. This marginalisation also shows that the intrinsic motivation function, \cref{eq:fepentropy}, is a functional of the complete posteriors or $\d(.,.,.|.)$.

In most publications on active inference the expected free energy in \cref{eq:fepentropy} is only part of what is referred to as the expected free energy. Usually, there is a second term measuring the relative entropy to an externally specified \textit{prior over future outcomes} (also called ``predictive distribution encoding goals'' \citealt{friston_active_2015}), i.e.\ a desired probability distribution $\p^d(\hs_\thT)$. The relative entropy term is formally given by:
\begin{align}
\label{eq:extrinsicvalue}
  \KL[\d(\hS_\thT|\ha_\thT)|| \p^d(\hS^d_\thT)]=\sum_{\hs_\thT} \d(\hs_\thT|\ha_\thT) \log \frac{\d(\hs_\thT|\ha_\thT)}{\p^d(\hs_\thT)}.
\end{align}
Clearly, this term will lead the agent to act such that the future distribution over sensor values is similar to the desired distribution. Since this term is used to encode extrinsic value for the agent, we mostly ignore it in this publication. It could included into any of the following intrinsic motivations.

In \citet{friston_active_curiosity_2017} yet another term, called ``negative novelty'' or ``ignorance'', occurs in the expected free energy. This term concerns the posterior distribution over parameter $\thetase$. It can be slightly generalised to refer to any subset of the parameters $\theta=(\thetase,\thetaeae,\thetae)$. We can write it as a conditional mutual information between future sensor values and parameters (the ``ignorance'' is the negative of this):
\begin{align}
\label{eq:infogain}
  \I_{\d}(\hS_\thT:\Theta|\ha_\thT)=\sum_{\hs_\thT} \d(\hs_\thT|\ha_\thT) \int \d(\theta|\hs_\thT,\ha_\thT) \log \frac{\d(\theta|\hs_\thT,\ha_\thT)}{\d(\theta)} \diff \theta.
\end{align}
This is identical to the information gain used in knowledge seeking agents. The necessary posteriors in the Bayesian case are $\q(\hs_\thT|\ha_\thT,sa_\pt,\xi)$, $\q(\theta|\hs_\thT,\ha_\thT,sa_\pt,\xi)$ and $\q(\theta|sa_\pt,\xi)$ with
\begin{align}
\label{eq:postpreds}
  \q(\hs_{t:\hT}|\ha_{t:\hT},sa_\pt,\xi)
&=  \int \sum_{\he_\pt} \q(\hs_\thT|\ha_\thT, \he_\tm,\theta) \q(\he_\pt, \theta|sa_\pt,\xi) \diff \theta 
\end{align}
a straightforward (if costly) marginalisation of the complete posterior. Just like previously for $\q(\he_\thT|\ha_\thT,sa_\pt,\xi)$, the marginalisation is greatly simplified in the variational case (see \cref{sec:sgapost} for a more explicit calculation). The integrals can be computed if using conjugate priors. The other two posteriors can be obtained via
\begin{align}
  \q(\theta|\hs_\thT,\ha_\thT,sa_\pt,\xi) = \frac{1}{\q(\hs_\thT|\ha_\thT,sa_\pt,\xi)} \sum_{\he_\allht} \q(\hs_\thT,\he_{0:\hT},\theta|\ha_\thT,sa_\pt,\xi) .
\end{align}
and
\begin{align}
  \q(\theta|sa_\pt,\xi) &= \q(\theta|\ha_\thT,sa_\pt,\xi)\\
  &= \sum_{\hs_\thT,\he_\allht} \q(\hs_\thT,\he_{0:\hT},\theta|\ha_\thT,sa_\pt,\xi) .
\end{align}
In the latter equation we used
\begin{align}
  \hA_\thT \ci \Theta | SA_\pt.
\end{align}
The marginalisations grow exponentially in computational cost with $\hT$. In this case, the variational approximation only reduces the necessary marginalisation over $\he_{\pt-1}$ to one over $\he_\tm$, but the marginalisation over future environment states $\he_\thT$ and sensor values $\hs_\thT$ remains the same since we use the exact predictive factor. In practice the time horizon into the future $\hT - t$ must then be chosen sufficiently short, so that marginalising out $\he_\thT$ and $\hS_\thT$ is feasible. Together with the variational approximation the required marginalisations over past and future are then constant over time which makes the implementation of agents with extended lifetimes possible.

The combination of the conditional entropy term and the information gain defines the (intrinsic part) of the action-value function of Friston's active inference (or free energy principle):
\begin{align}
\label{eq:fullfep}
  \mot^{FEP}(\d(.,.,.|.),\ha_\thT) = -\HS_{\d}(\hS_\thT|\hE_\thT)+\I_{\d}(\hS_\thT:\theta|\ha_\thT)
\end{align}
In the active inference literature this is usually approximated by a sum over the values at individual timesteps:
\begin{align}
\label{eq:timesumfep}
  \mot^{FEP}(\d(.,.,.|.),\ha_\thT) = \sum_{\tau = t}^\hT -\HS_{\d}(\hS_\tau|\hE_\tau)+\I_{\d}(\hS_\tau:\Theta|\ha_\thT).
\end{align}

\subsubsection{Free Energy Principle Specialised to \citet{friston_active_2015}}

Using \cref{appendix:translationTables}, we show how to get the action-value function of \citet[Eq. (9)]{friston_active_2015} in our framework.
In \citet{friston_active_2015}, the information gain of \cref{eq:infogain} is not included, but the extrinsic term of \cref{eq:extrinsicvalue} is. Furthermore, the sum over timesteps in \cref{eq:timesumfep} is used. This leads to the following expression:
\begin{align}
   \mot^{FEP}(\d(.,.,.|.),\ha_\thT) = \sum_{\tau = t}^\hT -\HS_{\d}(\hS_\tau|\hE_\tau) - \KL[\d(\hS_\tau|\ha_\thT)|| \p^d(\hS_\tau)].
\end{align}
If we plug in an approximate complete posterior, we get:
\begin{align}
\label{eq:fristonfep}
  \mot^{FEP}(\r(.,.,.|.),\ha_\thT) = \sum_{\tau = t}^\hT -\HS_{\r}(\hS_\tau|\hE_\tau) - \KL[\r(\hS_\tau|\ha_\thT)|| \p^d(\hS_\tau)].
\end{align}
with 
\begin{align}
  -\HS_{\r}(\hS_\tau|\hE_\tau) = \sum_{\he_\tau} \r(\he_\tau|\ha_\thT,\he_\tm,\phi) \sum_{\hs_\tau} \r(\hs_\tau|\he_\tau,\phi) \log \r(\hs_\tau|\he_\tau,\phi),
\end{align}
and
\begin{equation}
  \KL[\r(\hS_\tau|\ha_\thT)|| \p^d(\hS_\tau)] =\sum_{\hs_\tau} \r(\hs_\tau|\ha_\thT,\phi) \log \frac{\r(\hs_\tau|\ha_\thT,\phi)}{\p^d(\hs_\tau)}.
\end{equation} 
For the particular approximate posterior of \cref{eq:apposteriorxi}, with its factorisation into exact predictive and approximate posterior factor, the individual terms can be further rewritten. 
\begin{align}
  \r(\he_\tau|\ha_\thT,\he_\tm,\phi) &= \sum_{\hs_\thT,\he_{\tau+1:\hT}\he_{t:\tau-1}\he_{0:t-2}}\int\r(\hs_\thT,\he_{0:\hT},\theta|\ha_\thT,\phi) \diff \theta\\
  &=\sum_{\hs_\thT,\he_{\tau+1:\hT}\he_{t:\tau-1}\he_{0:t-2}}\int\q(\hs_\thT,\he_\thT|\ha_\thT, \he_\tm,\theta)\r(\he_\pt, \theta|\phi) \diff\theta\\
  &=\sum_{\hs_\thT,\he_{\tau+1:\hT}\he_{t:\tau-1}\he_{0:t-2}}\int\q(\hs_\thT,\he_\thT|\ha_\thT, \he_\tm,\theta)\prod_{r=0}^{t-1} \r(\he_r|\phir)  \prod_{i=1}^3 \r(\theta^i|\phi^i) \diff \theta\\ 
  &=\sum_{\he_{t:\tau-1}}\int\q(\he_{t:\tau-1}|\ha_\thT, \he_\tm,\thetaeae) \r(\he_\tm|\phitm)  \r(\thetaeae|\phieae) \diff \thetaeae\\
  &=\left(\sum_{\he_{t:\tau-1}}\int \prod_{r=t}^\tau \q(\he_r|\ha_r, \he_{r-1},\thetaeae) \r(\thetaeae|\phieae)\diff \thetaeae \right) \r(\he_\tm|\phitm). 
\end{align}

In \citet{friston_active_2015}, the environment dynamics $\q(\he_r|\ha_r, \he_{r-1},\thetaeae)$ are not inferred and are therefore not parameterised: 
\begin{align}
  \q(\he_r|\ha_r, \he_{r-1},\thetaeae) &= \q(\he_r|\ha_r, \he_{r-1})
\end{align}
and are set to the physical environment dynamics:
\begin{equation}
 \q(\he_r|\ha_r, \he_{r-1}) = \p(\he_r|\ha_r, \he_{r-1}).
\end{equation} 
This means the integral over $\thetaeae$ above is trivial and we get:
\begin{align}
  \r(\he_\tau|\ha_\thT,\he_\tm,\phi) &= \sum_{\he_{t:\tau-1}} \prod_{r=t}^\tau \q(\he_r|\ha_r, \he_{r-1}) \r(\he_\tm|\phitm)  \\
\end{align}
In the notation of \citet{friston_active_2015} (see \cref{sec:translationtable} for a translation table), we have 
\begin{align}
  \q(\he_r|\ha_r, \he_{r-1}) = \bB(\ha_r)_{\he_r \he_{r-1}}
\end{align}
where $\bB(\ha_r)$ is a matrix, and 
\begin{align}
  \r(\he_\tm|\phitm) = (\wideparen{s}_\tm)_{\he_\tm}
\end{align}
where $(\wideparen{s}_\tm)$ is a vector, so that 
\begin{align}
  \r(\he_\tau|\ha_\thT,\he_\tm,\phi)%
  &= (\bB(\ha_\tau) \cdots \bB(\ha_t) \cdot \wideparen{s}_\tm)_{\he_\tau}\\
  &=:(\wideparen{s}_\tau(\ha_\thT))_{\he_\tau}
\end{align}
Similarly, since the sensor dynamics in \citet{friston_active_2015} are also not inferred, we find
\begin{align}
  \r(\hs_\tau|\he_\tau,\phi) =\q(\hs_\tau|\he_\tau) =\p(\hs_\tau|\he_\tau).
\end{align}
\citeauthor{friston_active_2015} writes: 
\begin{align}
  \q(\hs_\tau|\he_\tau) =: \bA_{\hs_\tau \he_\tau}
\end{align}
with $\bA$ a matrix. So that,
\begin{align}
  \r(\hs_\tau|\ha_\thT,\phitm)&= \bA \cdot \wideparen{s}_\tau(\ha_\thT)\\
  &=:\wideparen{o}_\tau(\ha_\thT).
\end{align}
Then 
\begin{align}
  \HS_{\r}(\hS_\tau|\hE_\tau) = - \boldsymbol{1}\cdot (\bA \times \log \bA) \cdot \wideparen{s}_\tau(\ha_\thT)
\end{align}
where $\times$ is a Hadamard product and $\boldsymbol{1}$ is a vector of ones. Also, 
\begin{equation}
  \KL[\r(\hS_\tau|\ha_\thT)|| \p^d(\hS_\tau)] = \wideparen{o}_\tau(\ha_\thT) \cdot (\log \wideparen{o}_\tau(\ha_\thT) - \log \bC_\tau)
\end{equation} 
where $(\bC_\tau)_{\hs_\tau} = \p^d(\hs_\tau)$. Plugging these expressions into \cref{eq:fristonfep}, substituting $\ha_\thT \rightarrow \pi$, and comparing this to \citet[Eq. (9)]{friston_active_2015} shows that\footnote{There is a small typo in \citet[Eq. (9)]{friston_active_2015} where the time index of $\wideparen{s}_\tm$ in $(\wideparen{s}_\tau(\ha_\thT))= (\bB(\ha_\tau) \cdots \bB(\ha_t) \cdot \wideparen{s}_\tm)$ is given as $t$ instead of $\tm$.}:
\begin{align}
  \mot^{FEP}(\r(.,.,.|.),\pi)&= \boldsymbol{1}\cdot (\bA \times \log \bA) \cdot \wideparen{s}_\tau(\ha_\thT) - \wideparen{o}_\tau(\ha_\thT) \cdot (\log \wideparen{o}_\tau(\ha_\thT) - \log \bC_\tau)\\
  &=\bQ(\pi).
\end{align}
This verifies that our formulation of the action-value function specialises to the ``expected (negative) free energy'' $\bQ(\pi)$.

\subsubsection{Empowerment Maximisation}
\label{sec:empowerment}
Empowerment maximisation \citep{klyubin_empowerment_2005} is an intrinsic motivation that seeks to maximise the channel capacity from  sequences of the agent's actions into the subsequent sensor value. The agent, equipped with complete knowledge of the environment dynamics, can directly observe the environment state. If the environment is deterministic, an empowerment maximisation policy leads the agent to a state from which it can reach the highest number of future states within a preset number of actions. 

\citet{salge2014empowerment} provide a good overview of existing research on empowerment maximisation. A more recent study relates the intrinsic motivation to the essential dynamics of living systems, based on assumptions from autopoietic enactivism \cite{Guckelsberger2016b}. Several approximations have been proposed, along with experimental evaluations in complex state / action spaces. \citet{Salge2018} show how deterministic empowerment maximisation in a three-dimensional grid-world can be made more efficient by different modifications of UCT tree search. Three recent studies approximate stochastic empowerment and its maximisation via variational inference and deep neural networks, leveraging a variational bound on the mutual information proposed by \citet{barber2003algorithm}. \citet{mohamed_variational_2015} focus on a model-free approximation of open-loop empowerment, and \citet{gregor2016variational} propose two means to approximate closed-loop empowerment. While these two approaches  consider both applications in discrete and continuous state / action spaces, \citet{karl2017unsupervised} develop an open-loop, model-based approximation for the continuous domain specifically. The latter study also demonstrates how empowerment can yield good performance in established reinforcement learning benchmarks such as bipedal balancing in the absence of extrinsic rewards. In recent years, research on empowerment has particularly focused on applications in multi-agent systems. Coupled empowerment maximisation as a specific multi-agent policy has been proposed as intrinsic drive for either supportive or antagonistic behaviour in open-ended scenarios with sparse reward landscapes \cite{Guckelsberger2016a}. This theoretical investigation has then been backed up with empirical evaluations on supportive and adversarial video game characters \cite{Guckelsberger2016c,Guckelsberger2018}. Beyond virtual agents, the same policy has been proposed as a good heuristic to facilitate critical aspects of human-robot interaction, such as self-preservation, protection of the human partner, and response to human actions \cite{salge2017empowerment}. 

For empowerment, we select $\hTa=t+n$ and $\hT=t+n+m$, with $n\geq 0$ and $m\geq1$. This means the agent chooses $n+1$ actions which it expects to maximise the resulting $m$-step empowerment. The according action-value function is:
\begin{align}
  \mot^{EM}(\d(.,.,.|.),\ha_\thT) :&=  \max_{\d(\ha_{\hTa+1:\hT})} \; \I_{\d}(\hA_{\hTa+1:\hT}:\hS_\hT|\ha_\thTa) \\
  &=\max_{\d(\ha_{\hTa+1:\hT})} \; \sum_{\ha_{\hTa+1:\hT},\hs_\hT} \d(\ha_{\hTa+1:\hT}) \d(\hs_\hT|\ha_\thT) \log \frac{\d(\hs_\hT|\ha_\thT)}{\d(\hs_\hT|\ha_\thTa)}.  
\end{align}
Note that in the denominator of the fraction, the action sequence only runs to $\thTa$ and not to $\thT$ as in the numerator.

In the Bayesian case, the required posteriors are $\q(\hs_\hT|\ha_\thT,sa_\pt,\xi)$ (for each $\ha_{\hTa+1:\hT}$) and $\q(\hs_\hT|\ha_\thTa,sa_\pt,\xi)$. The former distribution is a further marginalisation over $\hs_{\tp:\hT-1}$ of $\q(\hs_\thT|\ha_\thT,sa_\pt,\xi)$. The variational approximation only helps getting $\q(\hs_\thT|\ha_\thT,sa_\pt,\xi)$, not the further marginalisation.
The latter distribution is obtained for a given $\q(\ha_{\hTa+1:\hT})$ from the former one via 
\begin{align}
  \q(\hs_\hT|\ha_\thTa,sa_\pt,\xi) &= \sum_{\ha_{\hTa+1:\hT}} \q(\hs_\hT,\ha_{\hTa+1:\hT}|\ha_\thTa,sa_\pt,\xi)\\
  &=\sum_{\ha_{\hTa+1:\hT}} \q(\hs_\hT|\ha_{\hTa+1:\hT},\ha_\thTa,sa_\pt,\xi) \q(\ha_{\hTa+1:\hT})
\end{align}
since the empowerment calculation imposes
\begin{align}
\q(\ha_{\hTa+1:\hT}|\ha_\thTa,sa_\pt,\xi)= \q(\ha_{\hTa+1:\hT}).
\end{align}

\subsubsection{Predictive Information Maximisation}
\label{sec:pim}
Predictive information maximisation, \citep{ay_predictive_2008}, is an intrinsic motivation that seeks to maximise the predictive information of the sensor process. Predictive information is the mutual information between past and future sensory signal, and has been proposed as a general measure of complexity of stochastic processes \citep{bialek1999predictive}. For applications in the literature see \citet{ay_information-driven_2012,martius_information_2013,martius_self-exploration_2014}. Also, see \citet{little_maximal_2013} for a comparison to entropy minimisation.

For predictive information, we select a half time horizon $k=\lfloor (\thT-t+1)/2 \rfloor$ where $k>0$ for predictive information to be defined (i.e.\ $\thT-t>0$). 
Then, we can define the expected mutual information between the next $m$ sensor values and the subsequent $m$ sensor values as the action-value function of predictive information maximisation. This is similar to the time-local predictive information in \citet{martius_information_2013}:
\begin{align}
\mot^{PI}(\d(.,.,.|.),\ha_\thT) :&= \I_{\d}(\hS_{t:t+k-1}:\hS_{t+k:t+2k-1}|\ha_\thT).
\end{align}
We omit writing out the conditional mutual information since it is defined in the usual way. Note that it is possible that $t+2k-1<\thT$ so that the action sequence $\ha_\thT$ might go beyond the evaluated sensor probabilities. This displacement leads to no problem since the sensor values do not depend on future actions. The posteriors needed are: $\q(\hs_{t:t+k-1}|\ha_\thT,sa_\pt,\xi)$, $\q(\hs_{t+k:t+2k-1}|\hs_{t:t+k-1},\ha_\thT,sa_\pt,\xi)$, and $\q(\hs_{t+k:t+2k-1}|\ha_\thT,sa_\pt,\xi)$. The first and the last are again marginalisations of $\q(\hs_\thT|\ha_\thT,sa_\pt,\xi)$ seen in \cref{eq:postpreds}. The second posterior is a fraction of such marginalisations.

\subsubsection{Knowledge Seeking}
\label{sec:ksa}
Knowledge seeking agents \citep{storck_reinforcement_1995,orseau_universal_2013} maximise the information gain with respect to a probability distribution over environments. The information gain we use here is the relative entropy between the belief over environments after actions and subsequent sensor values and the belief over environments (this is the KL-KSA of \citealt{orseau_universal_2013}, ``KL'' for Kullback-Leibler divergence). In our case the belief over environments can be identified with the posterior $\q(\theta|sa_\pt,\xi)$ since every $\theta=(\thetase,\thetaeae,\thetae)$ defines an environment. In principle, this can be extended to the posterior $\q(\xi|sa_\pt,\xi)$ over the hyperprior $\xi$, but we focus on $\theta$ here. This definition is more similar to the original one. Then, we define the knowledge seeking action-value function using the information gain of \cref{eq:infogain}:
\begin{align}
  \mot^{KSA}(\d(.,.,.|.),\ha_\thT) :&= \I_{\d}(\hS_\thT:\Theta|\ha_\thT).
\end{align}
We have discussed the necessary posteriors following \cref{eq:infogain}.

After this overview of some intrinsic motivations, we look at active inference. However, what should be clear is, that, in principle, both the posteriors needed for the intrinsic motivation function of the original active inference \citep{friston_active_2015} and the posteriors needed for alternative inferences overlap. This overlap shows that the other intrinsic motivations mentioned here also profit from variational inference approximations. There is also no indication that these intrinsic motivations cannot be used together with the next discussed active inference.

\section{Active Inference}

\label{sec:activeinference}
Now, we look at active inference. Note that this section is independent of the intrinsic motivation function underlying the action-value function $\hQ$. 

In the following we first look at and try to explain a slightly simplified version of the active inference in \citet{friston_active_2015}. Afterwards we also state the full version. 

As mentioned in the introduction, current active inference versions are formulated as an optimisation procedure that, at least at first sight, looks similar to the optimisation of a variational free energy familiar from variational inference. Recall that, in variational inference the parameters of a family of distributions are optimised to approximate an exact (Bayesian) posterior of a generative model. In the case we discussed in \cref{sec:approxpostandvi} the sought after exact posterior is the posterior factor of the generative model of \cref{sec:genmodel}. One of our questions about active inference is whether it is a straightforward application of variational inference to a posterior of some generative model. 
This would imply the existence of a generative model whose standard updating with past actions and sensor values leads to an optimal posterior distribution over future actions. Note that, this does not work with the generative model in of \cref{sec:genmodel} since the future actions there are independent of the past sensor values and actions. 
Given the appropriate generative model, it would then be natural to introduce it first and then apply a variational approximation similar to our procedure in \cref{sec:inference}.

We were not able to find in the literature or construct ourselves a generative model such that variational inference leads directly to the active inference as given in \citet{friston_active_2015}. Instead we present a generative model that contains a posterior whose variational approximation optimisation is very similar to the optimisation procedure of active inference. It is also closely related to the two-step action generation of first inferring the posterior and then selecting the optimal actions. This background provides some intuition for the particularities of active inference. 

One difference of the generative model used here is that its structure depends on the current time step in a systematic way. The previous generative model of \cref{sec:genmodel} had a time-invariant structure.

In \cref{sec:inference}, we showed how the generative model, together with either Bayesian or variational inference, can provide an agent with a set of complete posteriors. Each complete posterior is a conditional probability distribution over all currently unobserved variables ($\hS_\thT,\hE_\allt$) and parameters ($\Theta$ and more generally also $\Xi$) given past sensor values and actions $sa_\pt$ and a particular sequence of future actions $\ha_\thT$. Inference means updating the set of posteriors in response to observations $sa_\pt$. Active inference should then update the distribution over future actions in response to observations. This means the according posterior cannot be conditional on future action sequences like the complete posterior in \cref{eq:posteriorgfa2}. Since active inference promises belief or knowledge updating and action selection in one mechanism the posterior should also range over unobserved relevant variables like future sensor values, environment states, and parameters. This leads to the posterior of \cref{eq:posterior}:
\begin{align*}
  \q(\hs_\thT,\he_{0:\hT},\ha_\thT,\theta|sa_\pt,\xi). \tag{\ref{eq:posterior} \text{ revisited}}
\end{align*}
If this posterior has the right structure, then we can derive a future action distribution by marginalising:
\begin{align}
  \q(\ha_\thT|sa_\pt,\xi)=\sum_{\hs_\thT,\he_\allht} \int \q(\hs_\thT,\he_{0:\hT},\ha_\thT,\theta|sa_\pt,\xi) \diff \theta.
\end{align}
Actions can then be sampled from the distribution obtained by marginalising further to the next action only:
\begin{align}
\label{eq:actioninstantiation}
  \p(a_t|m_t):=\sum_{\ha_{t+1:\hT}} \q(\ha_\thT|sa_\pt,\xi).
\end{align}
This scheme could justifiably be called (non-variational) active inference since the future action distribution is directly obtained by updating the generative model. 

However, as we mentioned above, according to the generative model of \cref{fig:genmodel}, the distribution over future actions is independent of the past sensor values and actions:
\begin{align}
  \q(\hs_\thT,\he_{0:\hT},\ha_\thT,\theta|sa_\pt,\xi)=  \q(\hs_\thT,\he_{0:\hT},\theta|\ha_\thT,sa_\pt,\xi)\q(\ha_\thT)
\end{align}
since 
\begin{align}
  \q(\ha_\thT|sa_\pt,\xi)=  \q(\ha_\thT).
\end{align}
Therefore, we can never learn anything about future actions from past sensor values and actions using this model. In other words, if we intend to select the actions based on the past, we cannot uphold this independent model. The inferred actions must become dependent on the history and the generative model has to be changed for a scheme like the one sketched above to be successful.

\begin{figure}
\begin{center}
  \begin{tikzpicture}
    [->,>=stealth,auto,node distance=2cm,
    thick]
    \tikzset{
    hv/.style={to path={-| (\tikztotarget)}},
    vh/.style={to path={|- (\tikztotarget)}},
    }
    \tikzset{invi/.style={minimum width=0mm,inner sep=0mm,outer sep=0mm}}

    \node (ll2) [left of=al2,node distance=0.5cm,invi] {};
    \node (ll3) [left of=al3,node distance=0.5cm,invi] {};
    \node (ll1) [below of=al1,node distance=0.5cm,invi] {};

    \node (al2) [above of=al3, node distance=1cm] {$\Xieae$};
    \node (al3) [left of=th3] {$\Xie$};
    \node (al1) [below of=al3, node distance=3cm] {$\Xise$};

    \node (th2) [above of=th3, node distance=1cm] {$\Thetaeae$};
    \node (th3) [left of=el] {$\Thetae$};
    \node (th1) [below of=th3, node distance=3cm] {$\Thetase$};

    \node (el) [left of=e] {$\hE_0$};
    \node (sl) [below of=el, node distance=1cm] {$\hS_0$};
    \node (al) [below of=sl, node distance=1cm] {};
    \node (ml) [below of=al, node distance=1cm] {};

    \node (e) [] {$\hE_1$};
    \node (s) [below of=e, node distance=1cm] {$\hS_1$};   
    \node (a) [below of=s, node distance=1cm] {$\hA_1$};
    \node (th2') [above of=e', node distance=1cm] {};     
    \node (e') [right of=e, node distance=3cm] {$\hE_{2}$};
    \node (s') [below of=e', node distance=1cm] {$\hS_{2}$};
    \node (a') [below of=s', node distance=1cm] {$\hA_{2}$};
    \node (m') [below of=a', node distance=1cm] {};

    \node (th2r) [right of=th2'] {};       
    \node (er) [right of=e'] {$\hE_3$};
    \node (sr) [below of=er, node distance=1cm] {$\hS_3$};
    \node (ar) [below of=sr, node distance=1cm] {$\hA_3$};
    \node (mr) [below of=ar, node distance=1cm] {};

    \node (th2rr) [right of=th2r] {};    
    \node (err) [right of=er] {};
    \node (srr) [below of=err, node distance=1cm] {};
    \node (arr) [below of=srr, node distance=1cm] {$\phantom{\hA_4}$};
    \node (afrr) [below of=srr, node distance=.5cm,invi] {};

    \node (afdummy) [below of=sl, node distance=.75cm,invi] {};
    \node (afl) [right of=afdummy, node distance=.5cm,invi] {};
    \node (af) [right of=afl,invi] {};    
    \node (af') [right of=af, invi] {};
    \node (afr) [right of=af',] {};

    \node (afm) [left of=afr, node distance=1cm,invi] {};
    \node (afto3) [left of=afr, node distance=.5cm,invi] {};

    \node (dfl) [below of=afl, node distance=1cm,invi] {};
    \node (df) [right of=dfl,invi] {};    
    \node (df') [right of=df, invi] {};
    \node (dfr) [right of=df'] {};

    \node (dsdummy) [below of=a', node distance=.75cm,invi] {};
    \node (ds') [left of=dsdummy, node distance=.5cm,invi] {};
    \node (dsr) [right of=ds', invi] {};
    \node (dsrr) [right of=dsr, invi] {};
    \node (dsrra) [right of=dsrr,node distance=0.5cm, invi] {};

    \node (c0) [right of=th1, node distance=1cm,invi] {};
    \node (c1) [right of=c0,invi] {};
    \node (c2) [right of=c1,node distance=3cm,invi] {};
    \node (c3) [right of=c2,invi] {};
    \node (c4) [right of=c3,invi] {};
    \node (c3a) [right of=c2,node distance=1cm,invi] {};
    \node (c4a) [right of=c3,node distance=1cm,invi] {};
    \node (c5) [right of=c4a,invi] {};
    \node (xid') [below of=a',node distance=1.5cm,invi] {};
    \node (xidr) [right of=xid',invi] {};
    \node (xidrr) [right of=xidr,invi] {};    

    \path
      (al2) edge[-] (ll2)
      (ll2) edge[-,vh] (xidr)
      (al3) edge[-] (ll3)
      (ll3) edge[-,vh] (xidr)
      (al1) edge[-,vh] (xidr)
      (xidr) edge[-,dotted] (xidrr)
      (xid') edge (a')
      (xidr) edge (ar)
      ;

      \path
      (sl) edge[-] (afl)
      
      (afl) edge[-,vh] (ds')
      
      (s) edge[-] (af)
      (a) edge[-] (df)
      (af) edge[-] (df)

;
      \path
      (c4) edge[-,dotted,vh] (srr)
      ;
      \path
      (dsr) edge[-,line width=6pt,draw=white] (dsrra)
      (dsr) edge[-,dotted] (dsrra)

      (dsr) edge (ar)
      (ds') edge[-,line width=6pt,draw=white] (dsr)
      (ds') edge[-] (dsr)
      (ds') edge (a')
      (dsrr) edge[-,dotted] (arr)
;

      \path     
      (al3) edge (th3)
      (al2) edge (th2)
      (al1) edge (th1)

      (th3) edge (el)

      (th2) edge[hv] (e)

      ;
      \path
      (th1) edge[-,line width=6pt,draw=white] (c4a)
      ;
      \path
      (th1) edge[-] (c0)
      (c0) edge[vh] (sl)
      (c0) edge[-] (c1)
      (c1) edge[vh,line width=6pt,draw=white] (s)
      (c1) edge[vh] (s)
      ;

      \path
      
      (th2) edge[hv] (e')
      (th2) edge[hv] (er)
      (th2') edge[-,dotted] (th2rr)
      
      (c1) edge[-] (c2)
      (c2) edge[vh,line width=6pt,draw=white] (s')
      (c2) edge[vh] (s')
      (c2) edge[-] (c3)
      (c3) edge[vh,line width=6pt,draw=white] (sr)
      (c3) edge[vh] (sr)
      (c3) edge[-] (c4a)
      (c4a) edge[-,dotted,line width=6pt,draw=white] (c5) 
      (c4a) edge[-,dotted] (c5) 
;
      \path
      (el) edge node {} (e)
      (el) edge node {} (sl)      
      (e) edge node {} (s)

      (a) edge[bend left=45,line width=6pt,draw=white] node {} (e)   
      (a) edge[bend left=45] node {} (e)
      ;

      \path
      (e) edge node {} (e')
      
      (e') edge node {} (s')
      (a') edge[bend left=45,line width=6pt,draw=white] node {} (e')
      (a') edge[bend left=45] node {} (e')

      (e') edge (er)

      (ar) edge[bend left=45,line width=6pt,draw=white] node {} (er)
      (ar) edge[bend left=45] node {} (er)      
      (er) edge[-,dotted] (err)
      (er) edge (sr)
      ;

  \end{tikzpicture}
  \caption{Generative model including $\q(\ha_\thT|sa_\pt,\xi)$ at $t=2$ with $\hS\hA_{\prec 2}$ influencing future actions $\hA_{2:\hT}$. Note that, only future actions are dependent on past sensor values and actions, e.g.\ action $\hA_1$ has no incoming edges. The increased gap between time step $t=1$ and $t=2$ is to indicate that this time step is special in the model. For each time step $t$ there is an according model with the particular relation between past $\hS\hA_\pt$ and $\hA_\thT$ shifted accordingly.}
  \label{fig:activegenmodel}
\end{center}
\end{figure}
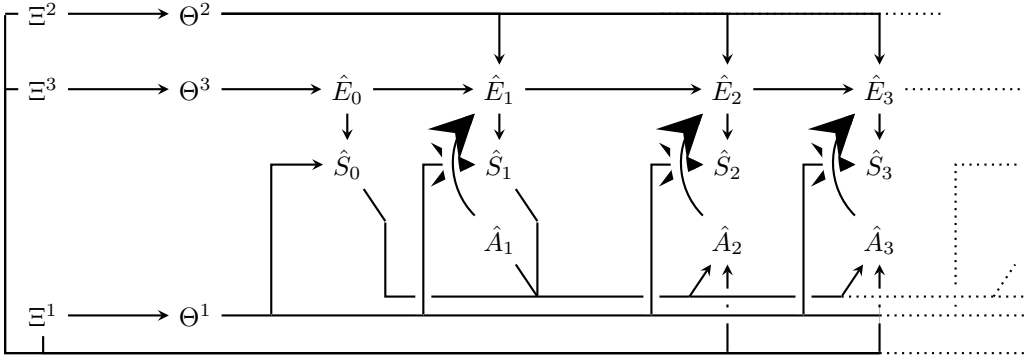

In \cref{sec:actionselection}, we have mentioned that the softmax policy based on a given action-value function $\hQ$ could be a desirable outcome of an active inference scheme such as the above. Thus, if we ended up with
\begin{align}
\label{eq:bayessoftmax}
  \q(\ha_\thT|sa_\pt,\xi)= \frac{1}{Z(\gamma,sa_\pt,\xi)} e^{\gamma \hQ(\ha_\thT,sa_\pt,\xi)}
\end{align}
as a result of some active inference process, that would be a viable solution. We can force this by building this conditional distribution directly into a new generative model. Note that this conditional distribution determines all future actions $\ha_\thT$ starting at time $t$ and not just the next action $\ha_t$. In the end however only the next action will be taken according to \cref{eq:actioninstantiation} and at time $t+1$ the action generation mechanism starts again, now with $\ha_{t+1:\hT}$ influenced by the new data $sa_t$ in addition to $sa_\pt$. So the model structure changes over time in this case with the dependency of actions on pasts $sa_\pt$ shifting together with each time-step. Keeping the rest of the previous Bayesian network structure intact we define that at each time $t$ the next action $\hA_t$ depends on past sensor values and actions $sa_\pt$ as well as on the hyperparameter $\xi$ (see \cref{fig:activegenmodel}):
\begin{align}
  \q(\hs_\thT,\he_{0:\hT},\ha_\thT,\theta|sa_\pt,\xi) := \q(\hs_\thT,\he_\thT|\ha_\thT,\he_\tm,\theta)\q(\ha_\thT|sa_\pt,\xi)\q(\theta,\he_\pt|sa_\pt,\xi).
\end{align}
On the right hand side we have the predictive and posterior factors left and right of the distribution over future actions. We define this conditional future action distribution to be the softmax of \cref{eq:bayessoftmax}. This means that the mechanism-generating future actions uses the Bayesian action-value function $\hQ(\ha_\thT,sa_\pt,\xi)$. The Bayesian action-value function depends on the complete posterior $\q(\hs_\thT,\he_\thT,\theta|\ha_\thT,sa_\pt,\xi)$ calculated using the old generative model of \cref{fig:genmodel} where actions do not not depend on past sensor values and actions. This is a complex construction with what amounts to Bayesian inference essentially happening within an edge (i.e.\ $\hS\hA_\pt \rightarrow \hA_\thT$) of a Bayesian network. However, logically there is no problem since the posterior $\q(\hs_\thT,\he_\thT,\theta|\ha_\thT,sa_\pt,\xi)$ for each $\ha_\thT$ to be well defined really only needs $sa_\pt$, $\xi$, and the model structure. Here we see the model structure as ``hard wired'' into the mechanism, since it is fixed for each time step $t$ from the beginning. 

We now approximate the posterior of \cref{eq:bayessoftmax} using variational inference. Like in \cref{sec:approxpostandvi} we do not approximate the predictive factor. Instead we only approximate the product of posterior factor $\q(\theta,\he_\pt|sa_\pt,\xi)$ and future action distribution $\q(\ha_\thT|sa_\pt,\xi)$. By construction these are two independent factors but with an eye to active inference which treats belief or knowledge updating and action generation together we also treat them together. For the approximation we again use the approximate posterio factor of \cref{eq:justappost} and combine it with a distribution over future actions $\r(\ha_\thT|\pi)$ parameterised by $\pi$:
\begin{align}
  \r(\ha_\thT,\he_\pt,\theta|\pi,\phi)&:=\r(\ha_\thT|\pi) \r(\he_\pt,\theta|\phi)\\
  &:=\r(\ha_\thT|\pi)\r(\he_\pt|\phipt)\r(\theta|\phi).
\end{align}
The variational free energy is then:
\begin{align}
\label{eq:activefreeenergy}
  \F[\pi,\phi,sa_\pt,\xi]:&= \sum_{\ha_\thT,\he_\pt} \int \r(\ha_\thT|\pi)\r(\he_\pt,\theta|\phi) \log \frac{\r(\ha_\thT|\pi)\r(\he_\pt,\theta|\phi)}{\q(s_\pt,\ha_\thT,\he_\pt,\theta|a_\pt,\xi)}\diff \theta\\
  &=\sum_{\ha_\thT,\he_\pt} \int \r(\ha_\thT|\pi)\r(\he_\pt,\theta|\phi) \log \frac{\r(\ha_\thT|\pi)\r(\he_\pt,\theta|\phi)}{\q(\ha_\thT|sa_\pt,\xi)\q(\he_\pt,\theta|sa_\pt,\xi)\q(s_\pt|a_\pt,\xi)}\diff \theta\\
  &=\F[\phi,sa_\pt,\xi]+\KL[\r(\hA_\thT|\pi)||\q(\hA_\thT|sa_\pt,\xi)].
\end{align}
Where $\F[\phi,sa_\pt,\xi]$ is the variational free energy of the (non-active) variational inference (see \cref{eq:fesimple}). Variational inference then minimises the above expression with respect to parameters $\phi$ and $\pi$:
\begin{align}
  \phis_{sa_\pt,\xi},\pi^*_{sa_\pt,\xi} :&=\argmin_{\phi,\pi} \F[\pi,\phi,sa_\pt,\xi]\\
  &=\argmin_{\phi} \F[\phi,sa_\pt,\xi] + \argmin_\pi \KL[\r(\hA_\thT|\pi)||\q(\hA_\thT|sa_\pt,\xi)]. \label{eq:splitmin}  
\end{align}
We see that the minimisation in this case separates into two minimisation problems. The first is just the variational inference of \cref{sec:approxpostandvi} and the second minimises the $\KL$-divergence between the parameterised action distribution $\r(\ha_\thT|\pi)$ and the softmax $\q(\ha_\thT|sa_\pt,\xi)$ of the Bayesian action-value function. 
It is instructive to look at this $\KL$-divergence term closer:
\begin{align}
   \KL[\r(\hA_\thT|\pi)||\q(\hA_\thT|sa_\pt,\xi)] &= -\HS_{\r}(\hA_\thT|\pi) - \sum_{\ha_\thT} \r(\ha_\thT|\pi) \log \q(\ha_\thT|sa_\pt,\xi)\\
   &= -\HS_{\r}(\hA_\thT|\pi) - \sum_{\ha_\thT} \r(\ha_\thT|\pi) \hQ(\ha_\thT,sa_\pt,\xi) + \log Z(\gamma,sa_\pt,\xi).
\end{align}
We see that the optimisation of $\pi$ leads towards high entropy distributions for which the expectation value of the action-value function $\hQ(\ha_\thT,\phi)$ is large. Action selection could then happen according to 
\begin{equation}
  \p(a_t|m_t):=\sum_{\ha_{t+1:T}} \r(\ha_\thT|\pi^*_{sa_\pt,\xi}).
\end{equation} 
So the described variational inference procedure, at least formally, leads to a useful result. However, this is not the active inference procedure of \citet{friston_active_2015}. As noted above the minimisation actually splits into two completely independent minimisations here. The result of the minimisation with respect to $\phi$ in \cref{eq:splitmin} is actually not used for action selection and since action selection is all that matters here is mere ornament. However, there is a way to make use of it. Recall that plugging $\phis_{sa_\pt,\xi}$ into the variational action-value function $\hQ(\ha_\thT,\phi)$ means that it approximates the Bayesian action value function (see \cref{eq:actionvalueapprox}). This means that if we define a softmax distribution $\r(\ha_\thT|\phi)$ of the variational action-value function parameterised by $\phi$ as:
\begin{align}
  \r(\ha_\thT|\phi)= \frac{1}{Z(\gamma,\phi)} e^{\gamma \hQ(\ha_\thT,\phi)}.
\end{align}
Then this approximates the softmax of the Bayesian action-value function:
\begin{align}
  \r(\ha_\thT|\phis_{sa_\pt,\xi}) \approx \q(\ha_\thT|sa_\pt,\xi).
\end{align}
Consequently, once we have obtained $\phis_{sa_\pt,\xi}$ from the first minimisation problem in \cref{eq:splitmin} we can plug it into $\r(\ha_\thT|\phi)$ and then minimise the $\KL$-divergence of $\r(\ha_\thT|\pi)$ to this distribution instead of the one to $\q(\ha_\thT|sa_\pt,\xi)$. In this way the result of the first could be reused for the second minimisation. This remains a two part action generation mechanism however. Active inference combines these two steps into one minimisation by replacing $\q(\ha_\thT|sa_\pt,\xi)$ in the variational free energy of \cref{eq:activefreeenergy} with $\r(\ha_\thT|\phi)$. Since $\r(\ha_\thT|\phi)$ thereby becomes part of the denominator it is also given the same symbol (in our case $\q$) as the generative model. So we define:
\begin{align}
  \q(\ha_\thT|\phi) :=\r(\ha_\thT|\phi). 
\end{align}
In this form the softmax $\q(\ha_\thT|\phi)$ is a cornerstone of active inference. In brief, it can be regarded as a prior over action sequences. To obtain purposeful behaviour it specifies prior assumptions about what sorts of actions an agent should take when its belief parameter takes value $\phi$.
Strictly speaking the expression resulting from the replacement $\q(\hA_\thT|sa_\pt,\xi) \rightarrow \q(\ha_\thT|\phi)$ in \cref{eq:activefreeenergy} is then not a variational free energy anymore since the variational parameters $\phi$ occur in both the numerator and the denominator. Nonetheless, this is the functional that is minimised in active inference as described in \citet{friston_active_2015}. So active inference is defined as the optimisation problem \citep[cmp.][Eq.(1)]{friston_active_2015}:
\begin{align}
    \phis_{sa_\pt,\xi},\pi^*_{sa_\pt,\xi} &=\argmin_{\phi,\pi}  \sum_{\ha_\thT,\he_\pt} \int \r(\ha_\thT|\pi)\r(\he_\pt,\theta|\phi) \log \frac{\r(\ha_\thT|\pi)\r(\he_\pt,\theta|\phi)}{\q(s_\pt,\ha_\thT,\he_\pt,\theta|\phi,a_\pt,\xi)}\diff \theta\\
    &=\argmin_{\phi,\pi} \left(\F[\phi,sa_\pt,\xi]+\KL[\r(\hA_\thT|\pi)||\q(\hA_\thT|\phi)]\right).
\end{align}
This minimisation does not split into the two independent parts anymore since both the future action distribution $\q(\hA_\thT|\phi)$ of the generative model and the approximate posterior factor in the variational free energy $\F[\phi,sa_\pt,\xi]$ are parameterised by $\phi$. This justifies the claim that active inference obtains both belief update and action selection through a single principle or optimisation.

Compared to \citet{friston_active_2015}, we have introduced a simplification of active inference. In the original text, additional distributions over $\gamma$ (with according random variable $\Gamma$) are introduced to the generative model as $\q(\gamma|\xig)$ (which is a fixed prior) and to the approximate posterior as $\r(\gamma|\phig)$. For the sake of completeness, we show the full equations as well. Since $\gamma$ is now part of the model, we write $\q(\ha_\thT|\gamma,\phi)$ instead of $\q(\ha_\thT|\phi)$. The basic procedure above stays the same. The active inference optimisation becomes: 

  \begin{align}
\label{eq:activeoptimisation}
    \begin{split}\phis_{sa_\pt,\xi},&\phigs_{sa_\pt,\xi},\pi^*_{sa_\pt,\xi}\\ 
    &=\argmin_{\phi,\phig,\pi}  \sum_{\ha_\thT,\he_\pt} \iint \r(\ha_\thT|\pi)\r(\gamma|\phig)\r(\he_\pt,\theta|\phi) \log \frac{\r(\ha_\thT|\pi)\r(\gamma|\phig)\r(\he_\pt,\theta|\phi)}{\q(s_\pt,\ha_\thT,\gamma,\he_\pt,\theta|\phi,a_\pt,\xi)} \diff\theta \diff \gamma.\end{split}
    \end{align}
Note that here, by construction, the denominator can be written as:
\begin{align}
  \q(s_\pt,\ha_\thT,\gamma,\he_\pt,\theta|\phi,a_\pt,\xi) = \q(\ha_\thT|\gamma,\phi) \q(\gamma|\phig) \q(\he_\pt,\theta|sa_\pt,\xi) \q(s_\pt|a_\pt,\xi).
\end{align}
Which allows us to write \cref{eq:activeoptimisation} with the original variational free energy again:
  \begin{align}
    \phis_{sa_\pt,\xi},&\phigs_{sa_\pt,\xi},\pi^*_{sa_\pt,\xi}
    &=\argmin_{\phi,\phig,\pi} \left(\F[\phi,sa_\pt,\xi] + \KL[\r(\hA_\thT,\Gamma|\pi,\phig)||\q(\hA_\thT,\Gamma|\phi,\xig)]\right).   
    \end{align}

\section{Applications and Limitations}
An application of the active inference described here to a simple maze task can be found in \citet{friston_active_2015}. Active inference using different forms of approximate posteriors can be found in \citet{friston_active_2016,friston_active_2016}. Here, \citet{friston_active_curiosity_2017} also includes a knowledge seeking term in addition to the conditional entropy term. In the universal reinforcement learning framework \citet{aslanides_universal_2017} also implement a knowledge seeking agent. These works can be quite directly translated into our framework. 

For applications of intrinsic motivations that are not so directly related to our framework see also the references in the according \cref{sec:empowerment,sec:pim,sec:ksa}.

A quantitative analysis of the limitations of the different approaches we discussed is beyond the scope of this publication. However, we can make a few observations that may help researchers interested in applying the discussed approaches.

Concerning the computation of the complete posterior by direct Bayesian methods is not feasible beyond the simplest of systems and even then only for very short time durations. As mentioned in the text it contains a sum over $|\shE|^t$ elements. If the time horizon into the future is $\hT-t$ then the predictive factor consists of $\shS^{\hT-t} \times \shE^{\hT-t} \times \shA^{\hT-t}$ entries. This means predicting far into the future is also not feasible. Therefore $\hT-t$ will usually have to be fixed to a small number. Methods that also approximate the predictive factor \citep[e.g.][]{friston_active_2016,friston_active_curiosity_2017} may be useful here. However, to our knowledge, their scalability has not been addressed yet. Since in these approaches the predictive factor is approximated in a similar way as the posterior factor here, we would expect that it is similar to the scalability of approximating the posterior factor.

Employing variational inference reduces the computational burden for obtaining a posterior factor considerably. The sum over all possible past environment histories (the $|\shE|^t$ elements) is approximated within the optimisation. Clearly, by employing variational inference we inherit all shortcomings of this method. As mentioned also in \citet{friston_active_2016} variational inference approximations are known to become overconfident i.e.\ the approximate posterior tends to ignore values with low probabilities \citep[see e.g.][]{bishop_pattern_2011}. In practice this can of course lead to poor decision making. Furthermore, the convergence of the optimisation to obtain the approximate posterior can also become slow. As time $t$ increases the necessary computations for each optimisation step in the widely used coordinate ascent variational inference algorithm \citep{blei_variational_2017} grow with $t^2$. Experiments suggest that the number of necessary optimisation steps also grows over time. At the moment, we do not know how fast but this may also lead to problems. A possible solution would be to introduce some form of forgetting such that the considered past does not grow forever. 

Ignoring the problem of obtaining a complete posterior, we still have to evaluate and select actions. Computing the information theoretic quantities needed for the mentioned intrinsic motivations and their induced action-value functions is also computationally expensive. In this case fixing the future time horizon $\hT-t$ can lead to constant computational requirements. These grow exponentially with the time horizon which makes large time horizons impossible without further approximations. Note that the action selection mechanisms discussed here also require the computation of the action-value functions for each of the future action sequences. 

Active inference is not a standard variational inference problem and therefore standard algorithms like the coordinate ascent variational inference may fail in this case. Other optimisation procedures like gradient descent may still work. As far as we know there have been no studies of the scalability of the active inference scheme up to now.

\section{Conclusion}
We have reconstructed the active inference approach of \citet{friston_active_2015} in in a formally consistent way. We started by disentangling the components of inference and action selection. This disentanglement has allowed us to also remove the variational inference completely and formulate the pure Bayesian knowledge updating for the generative model of \citet{friston_active_2015}. We have shown in \cref{sec:urlmodel} that a special case of this model is equivalent to a finite version of the model used by the Bayesian universal reinforcement agent \citep{hutter_universal_2005}. 
We then pointed out how to approximate the pure Bayesian knowledge updating with variational inference. To formalise the notion of intrinsic motivations within this framework, we have introduced intrinsic motivation functions that take complete posteriors and future actions as inputs. These induce action-value functions similar to those used in reinforcement learning. The action-value functions can then be used for both, the Bayesian and the variational agent, in standard deterministic or softmax action selection schemes.

Our analysis of the intrinsic motivations \emph{Expected Free Energy Maximisation}, \emph{Empowerment Maximisation}, \emph{Predictive Information Maximisation}, and \emph{Knowledge Seeking} indicates that there is significant common structure between the different approaches and it may be possible to combine them. 
At the time of writing, we have already made first steps towards using the present framework for a systematic quantitative analysis and comparison of the different intrinsic motivations. Eventually, such studies will shed more conclusive light on the computational requirements and emergent dynamics of different motivations. An investigation of the biological plausibility of different motivations might lead to different results and this is of equal interest.

Beyond the comparison of different intrinsic motivations within an active inference framework, the present work can thus contribute to investigations on the role of intrinsic motivations in living organisms. If biological plausibility of active inference can be upheld, and maintained for alternative intrinsic motivations, then experimental studies might be derived to test differentiating predictions. If active inference was key to cognitive phenomena such as consciousness, it would be interesting to see how the cognitive dynamics would be affected by alternative intrinsic motivations.

\section*{Conflict of Interest Statement}
CG, CS, SS, and DP declare no competing interests.
In accordance with Frontiers policy MB declares that he was employed by company Araya Incorporated, Tokyo, Japan.

\section*{Author Contributions}
MB, CG, CS, SS, and DP conceived of this study, discussed the concepts, revised the formal analysis, and wrote the article. MB contributed the initial formal analysis.

\section*{Funding}
CG is funded by EPSRC grant [EP/L015846/1] (IGGI). CS is funded by the EU Horizon 2020 programme under the Marie Sklodowska-Curie grant 705643. DP is funded in part by EC H2020-641321 socSMCs FET Proactive project.

\section*{Acknowledgments}
MB would like to thank Yen Yu for valuable discussions on active inference.

\begin{appendices}
\clearpage
\crefalias{section}{appsec}
\section{Posterior Factor}
\label{sec:postfactorBI}
Here we want to calculate the posterior factor $\q(\he_\pt, \theta|sa_\pt,\xi)$ of the complete posterior in \cref{eq:posteriorgfa2} without an approximation (i.e.\ as in direct, non-variational Bayesian inference).
\begin{align}
  \q(\he_\pt, \theta|sa_\pt,\xi)&= \frac{1}{\q(s_\pt|a_\pt,\xi)} \q(s_\pt,\he_\pt,\theta|a_\pt,\xi)\\
  &=\frac{1}{\q(s_\pt|a_\pt,\xi)} \q(s_\pt|\he_\pt,\thetase)\q(\he_\pt|a_\pt,\thetaeae,\thetae)\q(\theta|\xi)\\
  &=\frac{1}{\q(s_\pt|a_\pt,\xi)} \prod_{\tau=0}^t \q(s_\tau|\he_\tau,\thetase) \prod_{r=1}^t \q(\he_r|a_r,\he_{r-1},\thetaeae) \q(\he_0|\thetae) \prod_{i=1}^3\q(\theta^i|\xi^i).
\end{align}
We see that the numerator is given by the generative model. The denominator can be calulated according to:
\begin{align}
\label{eq:evidence}
  \q(s_\pt|a_\pt,\xi)& = \int_{\sTheta} \q(s_\pt|a_\pt,\theta) \q(\theta|\xi) \diff \theta\\
  &= \int_{\sTheta} \left( \sum_{\he_\pt}  \q(\he_0|\thetae) \prod_{\tau=0}^t \q(s_\tau|\he_\tau,\thetase) \prod_{r=1}^t\q(\he_r|a_r,\he_{r-1},\thetaeae) \right) \prod_{i=1}^3 \q(\theta^i|\xi^i) \diff \theta\\
  &=\sum_{\he_\pt} \int_{\sTheta} \q(\he_0|\thetae) \prod_{\tau=0}^t \q(s_\tau|\he_\tau,\thetase) \prod_{r=1}^t\q(\he_r|a_r,\he_{r-1},\thetaeae) \prod_{i=1}^3 \q(\theta^i|\xi^i) \diff \theta\\
  \begin{split}
    &=\sum_{\he_\pt} \left( \int \q(\he_0|\thetae) \q(\thetae|\xie)\diff\thetae \int \prod_{\tau=0}^t \q(s_\tau|\he_\tau,\thetase) \q(\thetase|\xise)\diff \thetase \right. \\
    &\phantom{=\sum_{\he_\pt} \left(\right.} \left. \times \int \prod_{r=1}^t\q(\he_r|a_r,\he_{r-1},\thetaeae) \q(\thetaeae|\xieae)\diff\thetaeae \right)
  \end{split}
\end{align}
The three integrals can be solved analytically if $\q(\theta^i|\xi^i)$ are chosen as conjugate priors to $\q(s_\tau|\he_\tau,\thetase),\q(\he_r|a_r,\he_{r-1},\thetaeae),\q(\he_0|\thetae)$ respectively. However, the sum is over $|\sE|^t$ terms and therefore untractable as time increases.

\clearpage
\section{Approximate Posterior Predictive Distribution}
\label{sec:sgapost}
Here, we calculate the (variational) approximate predictive posterior distribution of $\q(\hs_\thT|\ha_\thT ,sa_\pt,\xi)$ from a given approximate complete posterior. This expression plays a role in multiple intrinsic motivation functions like empowerment maximisation, predictive information maximisation, and knowledge seeking. For an arbitrary $\phi$ we have: 
\begin{align}
  \r(\hs_\thT|\ha_\thT ,\phi):&=\sum_{\he_\pt} \int \q(\hs_\thT|\ha_\thT,\he_\tm,\theta) \r(\he_\pt,\theta|\phi)  \diff \theta \\
  &=\sum_{\he_\tm} \int \q(\hs_\thT|\ha_\thT,\he_\tm,\theta) \r(\he_\tm,\theta|\phi)  \diff \theta \\
  &=\sum_{\he_\tm} \left(\int \q(\hs_\thT|\ha_\thT,\he_\tm,\theta) \prod_{i=1}^3 \r(\theta^i|\phi^i) \diff \theta \right) \r(\he_\tm|\phitm)   \\
  \begin{split}
    &=\sum_{\he_{\tm}}\left(\sum_{\he_\thT} \int \q(\hs_\thT|\he_\thT,\thetase) \r(\thetase|\phise) \diff \thetase \times \right.\\
    &\phantom{=\sum_{\he_{\tm}}\left(\sum_{\he_\thT}\right.}\left.\vphantom{\sum_{\he_\thT}}\times \int \q(\he_\thT|\ha_\thT,\he_\tm, \thetaeae) \r(\thetaeae|\phieae) \diff \thetaeae \right) \r(\he_\tm|\phitm) %
    \end{split}\\
  \begin{split}
    &=\sum_{\he_\tm} \left( \sum_{\he_\thT} \int \prod_{\tau=t}^{\hT} \q(\hs_\tau|\he_\tau,\thetase) \r(\thetase|\phise) \diff \thetase \times\right.\\
    &\phantom{=\sum_{\he_\tm}\left(\sum_{\he_\thT}\right.}\left.\vphantom{\sum_{\he_\thT}}\times \int \prod_{\tau=t}^{\hT} \q(\he_\tau|\ha_\tau,\he_{r-1}, \thetaeae) \r(\thetaeae|\phieae) \diff \thetaeae \right)\r(\he_\tm|\phitm)%
    \end{split}\\
   &=\sum_{\he_\tm} \sum_{\he_\thT} \r(\hs_\thT|\he_\thT,\phise) \r(\he_\thT|\ha_\thT,\he_\tm,\phieae) \r(\he_\tm|\phitm)
\end{align}
From first to second line we usually have to marginalize $\q(\he_\pt,\theta|sa_\pt,\xi)$ to $\q(\he_\tm,\theta|sa_\pt,\xi)$ with a sum over all $|\sE|^{t-1}$ possible environment histories $\he_{\pt-1}$. Using the approximate posterior, we can use $\r(\he_\tm|\phitm)$ directly without dealing with the intractable sum. From third to fourth line, $\r(\thetae|\phie)$ drops out since it can be integrated out (and its integral is equal to one). Note that during the optimisation \cref{eq:vi} $\r(\thetae|\phie)$ does play a role so it is not superfluous.%
From fifth to last line, we perform the integration over the parameters $\thetase$ and $\thetaeae$. These integrals can be calculated analytically if we choose the models $\r(\thetase|\phise)$ and $\r(\thetaeae|\phieae)$ as conjugate priors to $\q(s|e,\thetase)$ and $\q(e'|a',e,\thetaeae)$. Variational inference prediction of the next $n=\hT-t-1$ sensor values requires the sum and calculation of $|\shE|^n$ terms for $|\shS|^n$ possible futures. 

\clearpage
\section{Notation Translation Tables}
\label{appendix:translationTables}
A table to translate between our notation and the one used in \citet{friston_active_2015}. The translation is also valid in many cases for \citet{friston_active_2016,friston_active_learning_2016,friston_active_curiosity_2017}. Some of the parameters shown here only show up in the latter publications.

\label{sec:translationtable}
\vspace{.2cm}
\begin{tabularx}{\textwidth}{|L L X|}
\hline
  \text{This article} & \text{\citet{friston_active_2015}} & Note \\
  \hline
  e_t \in \sE &  & Actual environment states\\
  \he_t \in \shE & s_t\in S & Estimated/modelled environment states\\
  s_t \in \sS  & o_t \in \Omega & Actual/observed sensor or outcome values\\
  \hs_t \in \shS = \sS & o_t \in \Omega & Estimated/modelled (usually future) sensor or outcome values. Note that the index $\tau$ instead of $t$ often indicates an estimated future sensor value in \citet{friston_active_2015}. \\
  a_t \in \sA & a_t \in A & Actions\\
  \ha_t \in \shA =\sA & u_t \in U & Contemplated (usually future) actions\\
  m_t \in \sM & & Agent memory state\\
  \ha_\thT & \pi,\tilde{u} & $\pi$ and $\tilde{u}$ both uniquely specify future action sequences \\
  \theta & \theta & Generative model parameters \\
  \q(\hs|\he,\thetase)=\q(\hs|\he) & P(o|s)=\bA_{os} & Model sensor dynamics, not parameterised in \citet{friston_active_2015}, $\bA$ is a matrix representation\\
  \q(\he'|\ha',\he,\thetaeae)=\q(\he'|\ha',\he) & P(s'|s,u)= \bB(u)_{s's} & Model environment dynamics, not parameterised in \citet{friston_active_2015}, $\bB(u)$ is a matrix representation for each possible action $u$\\
  \q(\he_0|\thetae) & P(s_0|m) = \bD_{s_0} & Modelled initial environment state, not parameterised in \citet{friston_active_2015}, $\bD$ is a vector representation. Note, the parameter $m$ is a fixed hyperparameter\\  
  \xi = (\xise,\xieae,\xie) & m & Generative model hyperparam.\ or model parameter that subsumes all hyperparameters\\
  \xise &  & sensor dynamics hyperparam.\\
  \xieae &  & Environment dynamics hyperparam.\\
  \xie &  & Initial environment state hyperparam.\\
  \xig & (\alpha,\beta) & Precision hyperparam.\\
  (\phi,\phig) & \mu & Variational param.\\
  \phi^{E_\allht} & \wideparen{s} & Environment states variational param., \\
  \phitau & \wideparen{s}_\tau & for each timestep $\tau$\\
  \phise &  & Sensor dynamics variational param.\\
  \phieae &  & Environment dynamics variational param.\\
  \phie &  & Initial environment state variational param.\\
  \pi & \wideparen{\pi} & Future action sequence variational param.\\
  \phig & \wideparen{\gamma} &Precision variational param.\\
  \hQ(\ha_\thT,\phi) & \bQ(\pi)=\bQ(\tilde{u}|\pi) & Variational action-value function. The dependence of $\bQ(\tilde{u}|\pi)$ on $\wideparen{s}_t$ is omitted\\
  \p(s_\pet,e_\pet,a_\pt) & R(\tilde{o},\tilde{s},\tilde{a}) & Our physical environment corresponds to the generative process\\
    \q(\hs_\pet,\he_\pet,\ha_\thT,\gamma|a_\pt,\xi) & P(\tilde{o},\tilde{s},\tilde{u},\gamma|\tilde{a},m) & The generative model for active inference including $\gamma$ (which we mostly omit)\\
    \r(\he_\allht,\ha_\thT,\gamma|\pi,\phi,\phig) & Q(\tilde{s},\tilde{u},\gamma|\mu) & Approximate complete posterior for active inference\\ 
    \p^d(\hs_\tau) & P(o_\tau|m) & Prior over future outcomes.\\
  \hline
\end{tabularx}
Since our treatment is more general than that of \citet{friston_active_2015} and quite similar (though not identical) to the treatment in \citet{friston_active_2016,friston_active_learning_2016,friston_active_curiosity_2017} we also give the relations to variables in those publications. We hope this will help interested readers to understand the latter publications even if some aspects of those are different. A discussion of those differences is beyond the scope of the present article.

\begin{tabularx}{\textwidth}{|L L X|}
\hline
  \text{This article} & \text{\citet{friston_active_2016}} & Note \\
  \hline
  e_t \in \sE &  & Actual environment states\\
  \he_t \in \shE & s_t\in S & Estimated/modelled environment states\\
  s_t \in \sS  & o_t \in \Omega & Actual/observed sensor or outcome values\\
  \hs_t \in \shS = \sS & o_t \in \Omega & Estimated/modelled (usually future) sensor or outcome values. Note that the index $\tau$ instead of $t$ often indicates an estimated future sensor value in \citet{friston_active_2015}. \\
  a_t \in \sA & u_t \in A & Actions\\
  \ha_t \in \shA =\sA & u_t \in \varUpsilon & Contemplated (usually future) actions\\
  m_t \in \sM & & Agent memory state\\
  \ha_\allht & \pi, & action sequences \\
  \theta & \theta & Generative model parameters \\
  \thetase & \bA & Sensor dynamics param.\\
  \thetaeae & \bB & Environment dynamics param.\\
  \thetae & \bD & Initial environment state param.\\
  \xi & \eta & Generative model hyperparam.\ or model parameter that subsumes all hyperparameters\\\\
  \xise & a & sensor dynamics hyperparam.\\
  \xieae & b & Environment dynamics hyperparam.\\
  \xie & d & Initial environment state hyperparam.\\
  \xig & \beta & Precision hyperparam.\\
  (\phi,\phig) & \boldeta & Variational param.\\
  \phi^{E_\allht} &  \bs_{\allt} & Environment states variational param. \\
  \q(\he_\tau|\ha_\thT,a_{0:\tm},\phitau) & (\bs_\tau^\pi)_{\he_\tau} & For each sequence of actions and for each timestep there is a parameter $\bs_\tau^\pi$. Since a categorical distribution is used, the parameter is a vector of probabilities whose entry $\he_\tau$ is equal to the probability of $\he_\tau$ if we set $\shE=\{1,...,|\shE|\}$\\
  \phise & \ba & Sensor dynamics variational param.\\
  \phieae & \bb & Environment dynamics variational param.\\
  \phie & \bd & Initial environment state variational param.\\
  \pi & \bpi & Future action sequence variational param.\\
  \phig & \bbeta &Precision variational param.\\
  \hQ(\ha_\thT,\phi) & -\bG(\pi) & Variational action-value function. The dependence of $\bG(\pi)$ on $\bs_\allt^\pi$ is omitted\\
  \p(s_\pet,e_\pet,a_\pt) & R(\tilde{o},\tilde{s},\tilde{a}) & Our physical environment corresponds to the generative process\\
    \q(\hs_\pet,\he_\allht,\ha_\allht,\gamma,\theta,\xi) & P(\tilde{o},\tilde{s},\pi,\gamma,\bA,\bB,\bD|a,b,d,\beta) & The generative model for active inference\\
    \r(\he_\allht,\ha_\allht,\gamma,\theta|\pi,\phig,\phi) & Q(\tilde{s},\pi,\bA,\bB,\bD,\gamma|\bs^\pi_\allht,\bpi,\ba,\bb,\bd,\bbeta) & Approximate complete posterior for active inference\\ 
    \p^d(\hs_\tau) & P(o_\tau)=\sigma(\bU_\tau) & Prior over future outcomes.\\
  \hline
\end{tabularx}

\end{appendices}

\bibliographystyle{apalike} %
\bibliography{bibliography}

\begin{thebibliography}{}

\bibitem[Allen and Friston, 2016]{allen2016cognitivism}
Allen, M. and Friston, K.~J. (2016).
\newblock {From Cognitivism to Autopoiesis: Towards a Computational Framework
  for the Embodied Mind}.
\newblock {\em Synthese}, pages 1--24.

\bibitem[Aslanides et~al., 2017]{aslanides_universal_2017}
Aslanides, J., Leike, J., and Hutter, M. (2017).
\newblock {Universal Reinforcement Learning Algorithms: Survey and
  Experiments}.
\newblock In {\em Proceedings of the 26th International Joint Conference on
  Artificial Intelligence}, pages 1403--1410.

\bibitem[Attias, 1999]{attias_variational_1999}
Attias, H. (1999).
\newblock A {Variational} {Bayesian} {Framework} for {Graphical} {Models}.
\newblock In Solla, S., Leen, T., and M{\"u}ller, K., editors, {\em Proceedings
  Advances in Neural Information Processing Systems 12}, pages 209--215,
  Cambridge, MA, USA. MIT Press.

\bibitem[Attias, 2003]{attias_planning_2003}
Attias, H. (2003).
\newblock Planning by {Probabilistic} {Inference}.
\newblock In {\em Proceedings 9th International Workshop on Artificial
  Intelligence and Statistics}.

\bibitem[Ay et~al., 2012]{ay_information-driven_2012}
Ay, N., Bernigau, H., Der, R., and Prokopenko, M. (2012).
\newblock {Information-Driven Self-Organization: The Dynamical System Approach
  to Autonomous Robot Behavior}.
\newblock {\em Theory in Biosciences}, 131(3):161--179.

\bibitem[Ay et~al., 2008]{ay_predictive_2008}
Ay, N., Bertschinger, N., Der, R., Güttler, F., and Olbrich, E. (2008).
\newblock {Predictive Information and Explorative Behavior of Autonomous
  Robots}.
\newblock {\em The European Physical Journal B-Condensed Matter and Complex
  Systems}, 63(3):329--339.

\bibitem[Ay and L{\"o}hr, 2015]{ay_umwelt_2015}
Ay, N. and L{\"o}hr, W. (2015).
\newblock {The Umwelt of an Embodied Agent—a Measure-Theoretic Definition}.
\newblock {\em Theory in Biosciences}, 134(3-4):105--116.

\bibitem[Barber and Agakov, 2003]{barber2003algorithm}
Barber, D. and Agakov, F. (2003).
\newblock {The IM Algorithm: A Variational Approach to Information
  Maximization}.
\newblock In Thrun, S., Saul, L.~K., and Schölkopf, B., editors, {\em
  Proceedings Advances in Neural Information Processing Systems 16}, pages
  201--208. MIT Press.

\bibitem[Bialek and Tishby, 1999]{bialek1999predictive}
Bialek, W. and Tishby, N. (1999).
\newblock {Predictive Information}.
\newblock {\em arXiv preprint cond-mat/9902341}.

\bibitem[Bishop, 2011]{bishop_pattern_2011}
Bishop, C.~M. (2011).
\newblock {\em Pattern {Recognition} and {Machine} {Learning}}.
\newblock Information {Science} and {Statistics}. Springer, New York.

\bibitem[Blei et~al., 2017]{blei_variational_2017}
Blei, D.~M., Kucukelbir, A., and McAuliffe, J.~D. (2017).
\newblock Variational {Inference}: {A} {Review} for {Statisticians}.
\newblock {\em Journal of the American Statistical Association},
  112(518):859--877.

\bibitem[Botvinick and Toussaint, 2012]{botvinick_planning_2012}
Botvinick, M. and Toussaint, M. (2012).
\newblock {Planning as Inference}.
\newblock {\em Trends in Cognitive Sciences}, 16(10):485--488.

\bibitem[Buckley et~al., 2017]{buckley2017free}
Buckley, C.~L., Kim, C.~S., McGregor, S., and Seth, A.~K. (2017).
\newblock {The Free Energy Principle for Action and Perception: A Mathematical
  Review}.
\newblock {\em Journal of Mathematical Psychology}, pages 55--79.

\bibitem[Clark, 2015]{clark2015surfing}
Clark, A. (2015).
\newblock {\em {Surfing Uncertainty: Prediction, Action, and the Embodied
  Mind}}.
\newblock Oxford University Press.

\bibitem[Cover and Thomas, 2006]{cover_elements_2006}
Cover, T.~M. and Thomas, J.~A. (2006).
\newblock {\em {Elements of Information Theory}}.
\newblock Wiley-Interscience, Hoboken, {N.J.}

\bibitem[Dennett, 1991]{Dennett1991-DENCE}
Dennett, D.~C. (1991).
\newblock {\em Consciousness Explained}.
\newblock Penguin Books.

\bibitem[Doshi-Velez et~al., 2015]{doshi-velez_bayesian_2015}
Doshi-Velez, F., Pfau, D., Wood, F., and Roy, N. (2015).
\newblock Bayesian {Nonparametric} {Methods} for {Partially}-{Observable}
  {Reinforcement} {Learning}.
\newblock {\em IEEE Transactions on Pattern Analysis and Machine Intelligence},
  37(2):394--407.

\bibitem[Ellis and Wong, 2008]{ellis_learning_2008}
Ellis, B. and Wong, W.~H. (2008).
\newblock Learning {Causal} {Bayesian} {Network} {Structures} {From}
  {Experimental} {Data}.
\newblock {\em Journal of the American Statistical Association},
  103(482):778--789.

\bibitem[Fox and Tishby, 2016]{fox_minimum-information_2016}
Fox, R. and Tishby, N. (2016).
\newblock Minimum-information lgq control part ii: Retentive controllers.
\newblock In {\em 2016 IEEE 55th Conference on Decision and Control (CDC)},
  pages 5603--5609.

\bibitem[Friston, 2010]{friston2010free}
Friston, K. (2010).
\newblock The free-energy principle: A unified brain theory?
\newblock {\em Nature Reviews Neuroscience}, 11(2):127--138.

\bibitem[Friston, 2013a]{friston_consciousness_2013}
Friston, K. (2013a).
\newblock Consciousness and {Hierarchical} {Inference}.
\newblock {\em Neuropsychoanalysis}, 15(1):38--42.

\bibitem[Friston, 2013b]{friston_life_2013}
Friston, K. (2013b).
\newblock {Life as We Know It}.
\newblock {\em Journal of The Royal Society Interface}, 10(86).

\bibitem[Friston et~al., 2016a]{friston_active_learning_2016}
Friston, K., FitzGerald, T., Rigoli, F., Schwartenbeck, P., O'Doherty, J., and
  Pezzulo, G. (2016a).
\newblock {Active Inference and Learning}.
\newblock {\em Neuroscience \& Biobehavioral Reviews}, 68(Supplement
  C):862--879.

\bibitem[Friston et~al., 2016b]{friston_active_2016}
Friston, K., FitzGerald, T., Rigoli, F., Schwartenbeck, P., and Pezzulo, G.
  (2016b).
\newblock Active {Inference}: {A} {Process} {Theory}.
\newblock {\em Neural Computation}, 29(1):1--49.

\bibitem[Friston et~al., 2015]{friston_active_2015}
Friston, K., Rigoli, F., Ognibene, D., Mathys, C., Fitzgerald, T., and Pezzulo,
  G. (2015).
\newblock {Active Inference and Epistemic Value}.
\newblock {\em Cognitive Neuroscience}, 6(4):187--214.

\bibitem[Friston et~al., 2012]{friston_active_2012}
Friston, K., Samothrakis, S., and Montague, R. (2012).
\newblock {Active Inference and Agency: Optimal Control Without Cost
  Functions}.
\newblock {\em Biological Cybernetics}, 106(8-9):523--541.

\bibitem[Friston et~al., 2017a]{friston_active_curiosity_2017}
Friston, K.~J., Lin, M., Frith, C.~D., Pezzulo, G., Hobson, J.~A., and
  Ondobaka, S. (2017a).
\newblock Active {Inference}, {Curiosity} and {Insight}.
\newblock {\em Neural Computation}, 29(10):2633--2683.

\bibitem[Friston et~al., 2017b]{friston_graphical_2017}
Friston, K.~J., Parr, T., and de~Vries, B. (2017b).
\newblock {The Graphical Brain: Belief Propagation and Active Inference}.
\newblock {\em Network Neuroscience}, 1(4):381--414.

\bibitem[Froese and Ziemke, 2009]{froese_enactive_2009}
Froese, T. and Ziemke, T. (2009).
\newblock Enactive artificial intelligence: {Investigating} the systemic
  organization of life and mind.
\newblock {\em Artificial Intelligence}, 173(3--4):466--500.

\bibitem[Gregor et~al., 2016]{gregor2016variational}
Gregor, K., Rezende, D.~J., and Wierstra, D. (2016).
\newblock {Variational Intrinsic Control}.
\newblock {\em arXiv preprint arXiv:1611.07507}.

\bibitem[Guckelsberger and Salge, 2016a]{guckelsberger2016does}
Guckelsberger, C. and Salge, C. (2016a).
\newblock Does empowerment maximisation allow for enactive artificial agents?
\newblock In {\em Proceedings of the Fifteenth International Conference on the
  Synthesis and Simulation of Living Systems (Alife 2016)}, page~8. The MIT
  Press.

\bibitem[Guckelsberger and Salge, 2016b]{Guckelsberger2016b}
Guckelsberger, C. and Salge, C. (2016b).
\newblock {Does Empowerment Maximisation Allow for Enactive Artificial Agents?}
\newblock In {\em Proceedings 15th International Conference on Synthesis and
  Simulation of Living Systems (ALIFE)}.

\bibitem[Guckelsberger et~al., 2016a]{Guckelsberger2016c}
Guckelsberger, C., Salge, C., and Colton, S. (2016a).
\newblock {Intrinsically Motivated General Companion NPCs via Coupled
  Empowerment Maximisation}.
\newblock In {\em Proceedings Conference on Computational Intelligence in
  Games}.

\bibitem[Guckelsberger et~al., 2016b]{Guckelsberger2016a}
Guckelsberger, C., Salge, C., Saunders, R., and Colton, S. (2016b).
\newblock {Supportive and Antagonistic Behaviour in Distributed Computational
  Creativity via Coupled Empowerment Maximisation}.
\newblock In {\em Proceedings 7th International Conference on Computational
  Creativity}.

\bibitem[Guckelsberger et~al., 2018]{Guckelsberger2018}
Guckelsberger, C., Salge, C., and Togelius, J. (2018).
\newblock {New And Surprising Ways to be Mean: Adversarial NPCs with Coupled
  Empowerment Minimisation}.
\newblock In {\em Proceedings Conference on Computational Intelligence in
  Games}.

\bibitem[Hutter, 2005]{hutter_universal_2005}
Hutter, M. (2005).
\newblock {\em Universal {Artificial} {Intelligence}: {Sequential} {Decisions}
  {Based} on {Algorithmic} {Probability}}.
\newblock Texts in {Theoretical} {Computer} {Science}. {An} {EATCS} {Series}.
  Springer-Verlag, Berlin Heidelberg.

\bibitem[Karl et~al., 2017]{karl2017unsupervised}
Karl, M., Soelch, M., Becker-Ehmck, P., Benbouzid, D., van~der Smagt, P., and
  Bayer, J. (2017).
\newblock {Unsupervised Real-Time Control through Variational Empowerment}.
\newblock {\em arXiv preprint arXiv:1710.05101}.

\bibitem[Klyubin et~al., 2005]{klyubin_empowerment_2005}
Klyubin, A., Polani, D., and Nehaniv, C. (2005).
\newblock {Empowerment: A Universal Agent-Centric Measure of Control}.
\newblock In {\em The 2005 {IEEE} Congress on Evolutionary Computation, 2005},
  volume~1, pages 128--135.

\bibitem[Leike, 2016]{leike_nonparametric_2016}
Leike, J. (2016).
\newblock Nonparametric {General} {Reinforcement} {Learning}.
\newblock {\em arXiv:1611.08944 [cs]}.

\bibitem[Linson et~al., 2018]{linson_active_2018}
Linson, A., ~, A., Ramamoorthy, S., and Friston, K. (2018).
\newblock {The Active Inference Approach to Ecological Perception: General
  Information Dynamics for Natural and Artificial Embodied Cognition}.
\newblock {\em Frontiers in Robotics and AI}, 5:21.

\bibitem[Little and Sommer, 2013]{little_maximal_2013}
Little, D. Y.-J. and Sommer, F.~T. (2013).
\newblock Maximal mutual information, not minimal entropy, for escaping the
  “{Dark} {Room}”.
\newblock {\em Behavioral and Brain Sciences}, 36(3):220--221.

\bibitem[Lunn et~al., 2000]{lunn_winbugs_2000}
Lunn, D.~J., Thomas, A., Best, N., and Spiegelhalter, D. (2000).
\newblock {WinBUGS} - {A Bayesian Modelling Framework: {Concepts}, Structure,
  and Extensibility}.
\newblock {\em Statistics and Computing}, 10(4):325--337.

\bibitem[Manzotti and Chella, 2018]{10.3389/frobt.2018.00039}
Manzotti, R. and Chella, A. (2018).
\newblock Good old-fashioned artificial consciousness and the intermediate
  level fallacy.
\newblock {\em Frontiers in Robotics and AI}, 5:39.

\bibitem[Martius et~al., 2013]{martius_information_2013}
Martius, G., Der, R., and Ay, N. (2013).
\newblock Information {Driven} {Self}-{Organization} of {Complex} {Robotic}
  {Behaviors}.
\newblock {\em PLoS ONE}, 8(5).

\bibitem[Martius et~al., 2014]{martius_self-exploration_2014}
Martius, G., Jahn, L., Hauser, H., and Hafner, V.~V. (2014).
\newblock {Self-Exploration of the Stumpy Robot with Predictive Information
  Maximization}.
\newblock In del Pobil, A.~P., Chinellato, E., Martinez-Martin, E., Hallam, J.,
  Cervera, E., and Morales, A., editors, {\em From Animals to Animats 13: 13th
  International Conference on Simulation of Adaptive Behavior, SAB 2014,
  Castell{\'o}n, Spain, July 22-25, 2014. Proceedings}, pages 32--42. Springer.

\bibitem[Minka, 2001]{minka_expectation_2001}
Minka, T.~P. (2001).
\newblock Expectation {Propagation} for {Approximate} {Bayesian} {Inference}.
\newblock In {\em Proceedings of the {Seventeenth} {Conference} on
  {Uncertainty} in {Artificial} {Intelligence}}, {UAI}'01, pages 362--369, San
  Francisco. Morgan Kaufmann Publishers Inc.

\bibitem[Mohamed and Rezende, 2015]{mohamed_variational_2015}
Mohamed, S. and Rezende, D.~J. (2015).
\newblock {Variational Information Maximisation for Intrinsically Motivated
  Reinforcement Learning}.
\newblock In Cortes, C., Lawrence, N.~D., Lee, D.~D., Sugiyama, M., and
  Garnett, R., editors, {\em Proceedings Advances in Neural Information
  Processing Systems 28}, pages 2125--2133. Curran Associates, Inc.

\bibitem[Orseau et~al., 2013]{orseau_universal_2013}
Orseau, L., Lattimore, T., and Hutter, M. (2013).
\newblock {Universal Knowledge-Seeking Agents for Stochastic Environments}.
\newblock In Jain, S., Munos, R., Stephan, F., and Zeugmann, T., editors, {\em
  Algorithmic Learning Theory}, number 8139 in Lecture Notes in Computer
  Science, pages 158--172. Springer Berlin Heidelberg.

\bibitem[Ortega, 2011]{ortega_bayesian_2011}
Ortega, P.~A. (2011).
\newblock {Bayesian Causal Induction}.
\newblock {\em arXiv preprint arXiv:1111.0708}.

\bibitem[Ortega and Braun, 2010]{ortega_minimum_2010}
Ortega, P.~A. and Braun, D.~A. (2010).
\newblock {A Minimum Relative Entropy Principle for Learning and Acting}.
\newblock {\em Journal of Artificial Intelligence Research}, 38(1):475--511.

\bibitem[Ortega and Braun, 2014]{ortega_generalized_2014}
Ortega, P.~A. and Braun, D.~A. (2014).
\newblock {Generalized Thompson Sampling for Sequential Decision-Making and
  Causal Inference}.
\newblock {\em Complex Adaptive Systems Modeling}, 2:2.

\bibitem[Oudeyer and Kaplan, 2009]{oudeyer2009intrinsic}
Oudeyer, P.-Y. and Kaplan, F. (2009).
\newblock What is intrinsic motivation? a typology of computational approaches.
\newblock {\em Frontiers in Neurorobotics}, 1:6.

\bibitem[Oudeyer et~al., 2007]{oudeyer_intrinsic_2007}
Oudeyer, P.-Y., Kaplan, F., and Hafner, V.~V. (2007).
\newblock {Intrinsic Motivation Systems for Autonomous Mental Development}.
\newblock {\em {IEEE} Transactions on Evolutionary Computation},
  11(2):265--286.

\bibitem[Pearl, 2000]{pearl_causality_2000}
Pearl, J. (2000).
\newblock {\em {Causality: Models, Reasoning, and Inference}}.
\newblock Cambridge University Press.

\bibitem[Pfeifer et~al., 2005]{pfeifer2005}
Pfeifer, R., Iida, F., and Bongard, J. (2005).
\newblock {New Robotics: Design Principles for Intelligent Systems}.
\newblock {\em Artificial Life}, 11(1-2):99--120.

\bibitem[Ross and Pineau, 2008]{ross_model-based_2008}
Ross, S. and Pineau, J. (2008).
\newblock Model-{Based} {Bayesian} {Reinforcement} {Learning} in {Large}
  {Structured} {Domains}.
\newblock {\em Proceedings 24th Conference on Uncertainty in Artificial
  Intelligence}, 2008:476--483.

\bibitem[Ryan and Deci, 2000]{ryan_intrinsic_2000}
Ryan, R.~M. and Deci, E.~L. (2000).
\newblock Intrinsic and {Extrinsic} {Motivations}: {Classic} {Definitions} and
  {New} {Directions}.
\newblock {\em Contemporary Educational Psychology}, 25(1):54--67.

\bibitem[Salge et~al., 2014]{salge2014empowerment}
Salge, C., Glackin, C., and Polani, D. (2014).
\newblock {Empowerment--an Introduction}.
\newblock In {\em Guided Self-Organization: Inception}, pages 67--114.
  Springer.

\bibitem[Salge et~al., 2018]{Salge2018}
Salge, C., Guckelsberger, C., Canaan, R., and Mahlmann, T. (2018).
\newblock {Accelerating Empowerment Computation with UCT Tree Search}.
\newblock In {\em Proceedings Conference on Computational Intelligence in
  Games}. IEEE.

\bibitem[Salge and Polani, 2017]{salge2017empowerment}
Salge, C. and Polani, D. (2017).
\newblock {Empowerment as Replacement for the Three Laws of Robotics}.
\newblock {\em Frontiers in Robotics and AI}, 4:25.

\bibitem[Santucci et~al., 2013]{santucci_which_2013}
Santucci, V.~G., Baldassarre, G., and Mirolli, M. (2013).
\newblock {Which Is the Best Intrinsic Motivation Signal for Learning Multiple
  Skills?}
\newblock {\em Frontiers in Neurorobotics}, 7:22.

\bibitem[Schmidhuber, 2010]{schmidhuber_formal_2010}
Schmidhuber, J. (2010).
\newblock Formal {Theory} of {Creativity}, {Fun}, and {Intrinsic} {Motivation}
  (1990-2010).
\newblock {\em IEEE Transactions on Autonomous Mental Development},
  2(3):230--247.

\bibitem[Storck et~al., 1995]{storck_reinforcement_1995}
Storck, J., Hochreiter, S., and Schmidhuber, J. (1995).
\newblock {Reinforcement Driven Information Acquisition in Non-Deterministic
  Environments}.
\newblock In {\em Proceedings of the International Conference on Artificial
  Neural Networks}, volume~2, pages 159--164.

\bibitem[Sutton and Barto, 1998]{sutton_reinforcement_1998}
Sutton, R.~S. and Barto, A.~G. (1998).
\newblock {\em Reinforcement {Learning}: {An} {Introduction}}.
\newblock MIT Press.

\bibitem[Toussaint, 2009]{toussaint_probabilistic_2009}
Toussaint, M. (2009).
\newblock Probabilistic inference as a model of planned behavior.
\newblock {\em K{\"u}nstliche Intelligenz}, 3/09:23--29.

\bibitem[Vehtari et~al., 2014]{vehtari_expectation_2014}
Vehtari, A., Gelman, A., Sivula, T., Jylänki, P., Tran, D., Sahai, S.,
  Blomstedt, P., Cunningham, J.~P., Schiminovich, D., and Robert, C. (2014).
\newblock {Expectation Propagation as a Way of Life: A Framework for Bayesian
  Inference on Partitioned Data}.
\newblock {\em arXiv:1412.4869 [stat]}.

\bibitem[Wainwright and Jordan, 2007]{wainwright_graphical_2007}
Wainwright, M.~J. and Jordan, M.~I. (2007).
\newblock Graphical {Models}, {Exponential} {Families}, and {Variational}
  {Inference}.
\newblock {\em Foundations and Trends® in Machine Learning}, 1(1–2):1--305.

\bibitem[Winn and Bishop, 2005]{winn_variational_2005}
Winn, J. and Bishop, C.~M. (2005).
\newblock Variational {Message} {Passing}.
\newblock {\em Journal of Machine Learning Research}, 6(Apr):661--694.

\end{thebibliography}

\end{document}